\documentclass{article}

    \usepackage[preprint]{neurips_2025}

\usepackage{makecell}
\usepackage{hyperref}
\usepackage{float}
\usepackage{booktabs}
\usepackage{multirow}
\usepackage{wrapfig}
\usepackage{natbib}
\usepackage{microtype}
\usepackage{xcolor}         % colors
\usepackage{amsmath}
\usepackage{amssymb}
\usepackage{multirow}
\usepackage{multicol}
\usepackage{wrapfig}
\usepackage{tabularx}
\usepackage{booktabs}       % professional-quality tables
\usepackage{amsfonts}       % blackboard math symbols
\usepackage{graphicx}
\usepackage{nicefrac}  
\usepackage{pifont}
\usepackage{bbm}
\usepackage{colortbl}
\usepackage{longtable}
\usepackage{ulem}
\usepackage{listings}
\usepackage{xspace}
\usepackage{fancyvrb}
\usepackage{xcolor}
\usepackage[most]{tcolorbox}
\usepackage[utf8]{inputenc} % allow utf-8 input
\usepackage[T1]{fontenc}    % use 8-bit T1 fonts
\usepackage{hyperref}       % hyperlinks
\usepackage{url}            % simple URL typesetting
\usepackage{booktabs}       % professional-quality tables
\usepackage{amsfonts}       % blackboard math symbols
\usepackage{nicefrac}       % compact symbols for 1/2, etc.
\usepackage{microtype}      % microtypography
\usepackage{xcolor}         % colors

\usepackage{amsmath} 
\usepackage{enumitem} 
\usepackage{fontawesome5}  % 或 \usepackage{fontawesome}
\usepackage[utf8]{inputenc}
\newcommand{\github}{\raisebox{-1.5pt}{\includegraphics[height=1.05em]{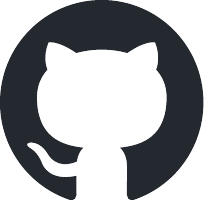}}\xspace}
\newcommand{\huggingface}{\raisebox{-1.5pt}{\includegraphics[height=1.05em]{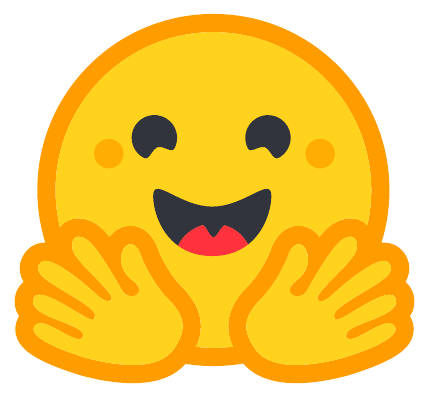}}\xspace}
\usepackage{ulem}
\usepackage{color}

\usepackage{subcaption}
\usepackage{xcolor}
\usepackage{pifont}
\usepackage{fontawesome5}

\title{Reinforcing Spatial Reasoning in Vision-Language Models with Interwoven Thinking and Visual Drawing}
\author{  Junfei Wu$^{123}$\thanks{Equal Contribution. $^\dagger$\text{Corresponding Author.}}, Jian Guan$^{3*}$, Kaituo Feng$^{4}$, Qiang Liu$^{12}$, Shu Wu$^{12\dagger}$, Liang Wang$^{12}$, \\ \textbf{ Wei Wu$^{3\dagger}$, Tieniu Tan$^{125}$} \\ 
  $^1$Institute of Automation, Chinese Academy of Sciences. \\
  $^2$University of Chinese Academy of Sciences.
  $^3$Ant Group. \\$^4$CUHK MMLab. $^5$Nanjing University.\\
\texttt{junfei.wu@cripac.ia.ac.cn, shu.wu@nlpr.ia.ac.cn,}\\
\texttt{\{jianguanthu, wuwei19850318\}@gmail.com}\\\\
{\small \github \texttt{\textbf{Code:}} \url{https://github.com/AntResearchNLP/ViLaSR}}\\
{\small \huggingface \texttt{\textbf{Model:}} \url{https://huggingface.co/AntResearchNLP/ViLaSR}}}
\begin{document}

\maketitle

% {\large \faGithub} \href{https://github.com/AntResearchNLP/ViLaSR}{Github: AntResearchNLP/ViLaSR}\\[0.2em]
% {\large \faHuggingFace} \href{https://huggingface.co/Hyperwjf/ViLaSR}

\begin{abstract}

As textual reasoning with large language models (LLMs) has advanced significantly, there has been growing interest in enhancing the multimodal reasoning capabilities of large vision-language models (LVLMs). However, existing methods primarily approach multimodal reasoning in a straightforward, text-centric manner, where both reasoning and answer derivation are conducted purely through text, with the only difference being the presence of multimodal input. As a result, these methods often encounter fundamental limitations in spatial reasoning tasks that demand precise geometric understanding and continuous spatial tracking\textemdash capabilities that humans achieve through mental visualization and manipulation. To address the limitations, we propose drawing to reason in space, a novel paradigm that enables LVLMs to reason through elementary drawing operations in the visual space. By equipping models with basic drawing operations, including annotating bounding boxes and drawing auxiliary lines, we empower them to express and analyze spatial relationships through direct visual manipulation, meanwhile avoiding the performance ceiling imposed by specialized perception tools in previous tool-integrated reasoning approaches. To cultivate this capability, we develop a three-stage training framework: cold-start training with synthetic data to establish basic drawing abilities, reflective rejection sampling to enhance self-reflection behaviors, and reinforcement learning to directly optimize for target rewards. Extensive experiments demonstrate that our model, named \textbf{\textsc{ViLaSR}}, consistently outperforms existing methods across diverse spatial reasoning benchmarks, involving maze navigation, static spatial reasoning, video-based reasoning, and multi-view-based reasoning tasks, with an average improvement of 18.4\%. Ablation studies reveal the critical role of each training stage, where reflective rejection sampling strengthens the model's self-correction capabilities, and reinforcement learning effectively unlocks its reasoning potential.

% Extensive experimental results demonstrate that our model consistently achieves state-of-the-art performance across various spatial reasoning benchmarks, including temporal-spatial, static spatial, and maze reasoning tasks. Notably, our model attains a performance rate of 41.1\% on the VSI-Bench, outperforming open-source models and rivaling advanced proprietary models. These findings underscore the robustness and generalization capabilities of our approach.

\end{abstract}

\section{Introduction}
Large language models (LLMs) have exhibited remarkable reasoning capabilities in complex tasks such as mathematical problem-solving~\cite{cot,lightman2023let} and code generation~\cite{chen2021evaluatinglargelanguagemodels}, particularly through the ``slow thinking'' paradigm exemplified by OpenAI o1~\cite{jaech2024openai} {and DeepSeek R1~\cite{deepseekai2025deepseekr1incentivizingreasoningcapability}}, which enables extended reasoning with in-depth self-reflection. {Encouraged by the success,  a growing body of research is now aiming to adapt similar techniques to large vision-language models (LVLMs) to enhance their capabilities in image and video reasoning~\citep{yang2025r1, deng2025openvlthinker, peng2025skywork, feng2025video}.}

While current LVLMs excel at basic visual perception tasks such as object detection~\cite{du2022learning} and narrative understanding~\cite{maaz2024video,cheng2025scalingvideolanguagemodels10k}, they often struggle with spatial reasoning~\cite{wang2024isapicture,yang2024think}, which requires understanding spatial relationships among objects and tracking their dynamic evolution~\cite{song2025robospatial}\textemdash capabilities that are crucial for real-world applications such as robotics~\cite{10611477} and augmented reality~\cite{Grauman2021Ego4DAT}. Indeed, even when such deep visual understanding is required, LVLMs still rely solely on text-based reasoning~\cite{yang2025r1}, assuming that visual information can be perfectly translated into textual semantic space~\cite{huh2024position}. Unfortunately, such translation is inherently challenging~\cite{li2024multimodal}: spatial details are inevitably lost when converting visual information to text, and describing dynamic changes of object positions becomes prohibitively complex in textual space (see examples of GPT-4o in Figure~\ref{fig:paradigm}). Taking inspiration from human cognition, where spatial reasoning relies on mental visualization and dynamic manipulation~\cite{burgess2008spatial}, we advocate a vision-centric reasoning paradigm where models actively edit and re-encode visual information at each reasoning step, dynamically supplementing spatial details and relationships. This ``thinking with images'' approach, while validated by OpenAI o3~\cite{openai2025thinking}, remains underexplored in open-source research.

In parallel with our work, recent studies have begun to integrate visual tools to enable vision-centric reasoning~\cite{cheng2024least,qi2025cogcom,hu2024visual,li2025imagine}. 
% begun to address the challenges of vision-centric reasoning by integrating visual tools into the reasoning process
However, these approaches exhibit limitations along two key dimensions. First, the reasoning capabilities of LVLMs are constrained by black-box perception tools (e.g., grounding and OCR systems), resulting in not only fixed perception capabilities but also fragmented reasoning composed of disconnected tool invocations that undermine coherence and holistic planning. Second, these methods heavily draw upon reasoning data curated based on human priors, which often exhibit oversimplified logic compared to the complexity of spatial reasoning tasks, with problem decomposition and tool invocation interleaved in a simplistic and linear fashion~\cite{cheng2024least,qi2025cogcom}. Such adherence to prescribed reasoning patterns prevents LVLMs from developing the capacity for critical reflection on tool outputs—a capability that has proven crucial for advanced reasoning~\cite{openai2024o1preview}. These fundamental limitations call for a more flexible and intrinsic approach to vision-centric reasoning.
In light of these challenges, we propose ``drawing to reason in space,'' a novel, versatile reasoning paradigm that empowers LVLMs to reason through elementary drawing operations, as exemplified in Figure~\ref{fig:paradigm}. Through simple yet powerful operations, including bounding boxes for object localization and auxiliary lines for relationship analysis, this paradigm enables direct visual interaction for spatial problem-solving, mirroring human behavior while avoiding the pitfalls of dependence on external perception tools. Based on this paradigm, we develop \textbf{\textsc{ViLaSR}}, a \underline{\textbf{\textsc{Vi}}}sion-\underline{\textbf{\textsc{La}}}nguage model that achieves sophisticated \underline{\textsc{\textbf{S}}}patial \underline{\textsc{\textbf{R}}}easoning through interwoven thinking and visual drawing. 
% \textbf{\textsc{ViLaSR}}, a VLM that achieves sophisticated \underline{\textsc{\textbf{Spa}}}tial reasoning through interwoven visual d\underline{\textsc{\textbf{r}}}awing and thin\underline{\textsc{\textbf{k}}}ing in space. 
To realize this vision, we develop a three-stage training framework~(Figure~\ref{fig:ViLaSR_train}). First, we introduce cold-start training with synthetic data to establish basic visual drawing abilities. Second, we design a reflective rejection sampling mechanism that selectively reinforces reasoning paths demonstrating both correct answers and self-correction behaviors, enabling models to revise their visual operations based on intermediate results. Finally, we employ reinforcement learning~(RL) with carefully designed rewards that balance answer correctness and reasoning format, incentivizing both accurate spatial understanding and coherent visual thinking processes. 
% This paradigm enables direct \textbf{\textsc{Spa}}tial reasoning in the visual space that interweaves d\textbf{\textsc{r}}awing and thin\textbf{\textsc{k}}ing~(\textbf{\textsc{ViLaSR}}). 
% Through simple yet powerful operations like bounding boxes for object localization and auxiliary lines for relationship analysis, \textsc{ViLaSR} mirrors human behavior in spatial problem-solving through direct visual interaction while avoiding the pitfalls of dependence on external perception tools. 
% This minimalist design not only preserves the richness of visual information but also empowers models to develop genuine reasoning capabilities through direct visual interaction. 
% Our framework differs fundamentally from previous approaches by focusing on developing models' inherent reasoning capabilities rather than relying on specialized perception models or prescribed reasoning paths.

Extensive experiments across five diverse spatial reasoning benchmarks demonstrate the effectiveness of \textsc{ViLaSR}, which achieves substantial improvements over strong baselines by 18.4\% on average, involving challenging scenarios that require sequential spatial planning (e.g., maze navigation), temporal relationship tracking (e.g., video-based reasoning), and information integration from multiple perspectives. Ablation studies reveal the critical role of each training stage. Particularly, reflective rejection sampling significantly enhances the model's self-correction capabilities, with the frequency of reflection behaviors doubling compared to models trained without this intermediate stage. Further inference-time scaling experiments reveal that each training stage progressively enhances the model's reasoning potential, with the final RL optimization effectively consolidating multi-attempt capabilities while maintaining strong single-attempt performance, significantly narrowing the performance gap compared to earlier training stages.

 %, highlighting the effectiveness of our training framework in cultivating sophisticated reasoning behaviors.

% Experiments on diverse spatial reasoning benchmarks demonstrate the effectiveness of \textsc{ViLaSR} across maze navigation, static image-based spatial understanding, and video spatial reasoning tasks. \textsc{ViLaSR} consistently outperforms previous approaches by significant margins (x\% - y\% absolute improvements). Notably, through drawing operations, our model exhibits strong performance in challenging scenarios that require dynamic spatial tracking (e.g., planning paths in mazes) or multi-frame relationship inference (e.g., analyzing object interactions across video frames). Ablation studies further validate the importance of each training stage and reveal that the model develops increasingly sophisticated reasoning patterns, from basic spatial annotation to complex self-correction behaviors. Particularly, we observe that the reflective rejection sampling stage significantly enhances the emergence of reflective behaviors during RL optimization, leading to substantial performance gains (z\% improvement) compared to models trained without this intermediate stage.

In summary, our contributions are threefold: 

% \begin{itemize}[leftmargin=10pt]
I. We propose drawing to reason in space, a novel reasoning paradigm that enables LVLMs to perform spatial reasoning through interpretable visual operations;
    
II. We develop a principled training framework that effectively cultivates models' visual reasoning capabilities through progressive stages;

III. We demonstrate state-of-the-art performance across multiple spatial reasoning benchmarks and provide insights into the development of visual reasoning abilities in LVLMs. 

\section{Related works}

% \subsection{Multimodal Reasoning with Reinforcement Learning}
% code interpreters~\citep{gao2023pal},
\subsection{Reasoning in LLMs and LVLMs}
Instilling advanced reasoning capabilities within LLMs and LVLMs remains a significant challenge. Current approaches can be broadly categorized into three directions: \textbf{(1) Prompt engineering:} crafting prompts to elicit latent reasoning capabilities in pre-trained models~\citep{cot,zhang2024multimodal} and enable tool usage to complement model capabilities with external knowledge~\cite{asai2023self,guan2024amor} or specialized functionalities~\citep{shen2023hugginggpt,gupta2023visual,suris2023vipergpt,yang2023mm}. These prompt-based approaches heavily depend on models' instruction-following capabilities and often suffer from prompt sensitivity. \textbf{(2) Supervised fine-tuning:} developing tailored datasets to enhance specific reasoning capabilities~\citep{yu2023metamath,wen2024codeplan,zhao2025promptcotsynthesizingolympiadlevelproblems} and training models to effectively utilize external tools~\citep{schick2024toolformer,qin2023toolllm,guan2024amor,cheng2024least,shao2024visual,qi2025cogcom}. Such approaches are inherently bounded by the quality and scale of training data, limiting their potential for advanced reasoning. \textbf{(3) Reinforcement learning:} engineering reward mechanisms to incentivize desired reasoning patterns~\citep{le2022coderl, shen2023pangu,deepseekai2025deepseekr1incentivizingreasoningcapability} and optimize tool usage strategies~\citep{li2025torlscalingtoolintegratedrl}. While promising in textual domains, RL approaches for tool-augmented reasoning have yet to be fully investigated in multimodal scenarios. Current LVLMs either confine their reasoning to textual form~\cite{yang2025r1}, or rely on specialized perception tools, being constrained by tool capabilities~\cite{cheng2024least,qi2025cogcom} and lacking reflective reasoning~\cite{shao2024visual}. In contrast, we pioneer a staged training recipe that enables models to both express and reflect upon their spatial understanding through iterative drawing operations.

\begin{figure}[!t]
  \centering
  \includegraphics[width=\linewidth]{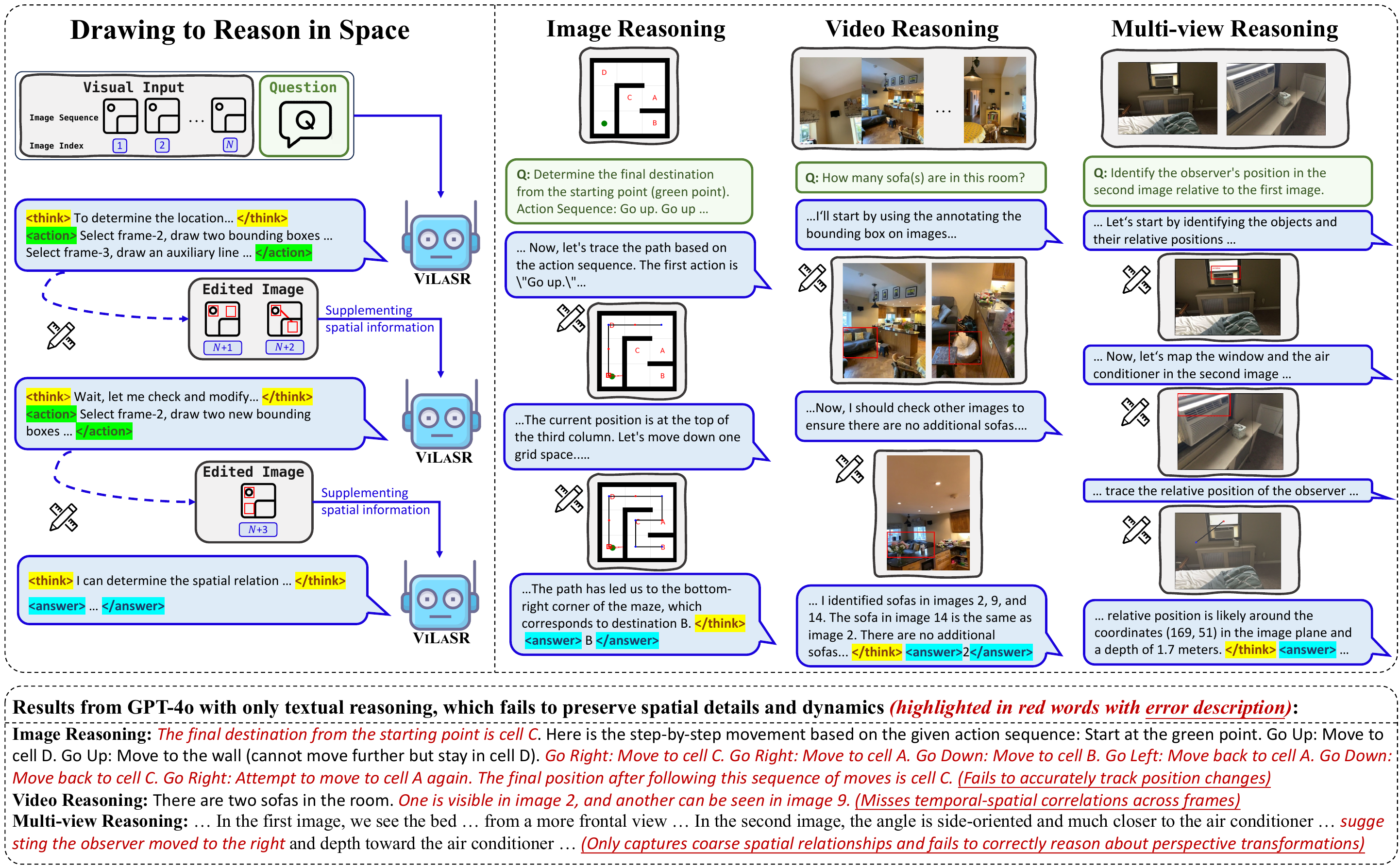}
  \vspace{-15pt}
  \caption{\textbf{Top left: }Overview of the ``drawing to reason in space'' paradigm, which enables visual reasoning through iterative thinking and drawing operations. \textbf{Top right: }Examples across three spatial reasoning tasks, demonstrating how \textsc{ViLaSR} decomposes complex problems into interpretable visual reasoning steps. \textbf{Bottom:} Corresponding results from GPT-4o with only textual reasoning.}
  % \gnote{update the model name in the figure. Can we find examples with reflection?} \gnote{can we add results from models with only textual reasoning (GPT-4o) and methods that first generate captions using VLMs (e.g. Qwen2.5-vl) and then use textual reasoning with LLMs (e.g., ds-r1), in order to illustrate the necessity of ``thinking with images.''}}
  \label{fig:paradigm}
  \vspace{-15pt}
\end{figure}

\subsection{Visual spatial reasoning}
Visual spatial reasoning, as a crucial component of multimodal intelligence, extends beyond general visual perception to encompass two particularly challenging fundamental capabilities~\cite{gardner2011frames,zhang2025scalingbeyondadvancingspatial}: relational reasoning, which involves understanding distances, directions, and spatial common sense between objects~\cite{liu-etal-2023-visual,song2025robospatial}; and perspective transformation, which requires holding and manipulating spatial relationships~\cite{mcafoose2009exploring,yang2024think}. Despite the remarkable progress of current LVLMs in basic visual tasks~\cite{hurst2024gpt,lin2024vila,liu2024improved,cheng2025scalingvideolanguagemodels10k}, numerous benchmarks have revealed significant challenges in spatial reasoning~\cite{mirzaee-etal-2021-spartqa,kamath2023s,yamada2023evaluating,li2024topviewrsvisionlanguagemodelstopview,yang2024think}. This limitation is partly attributed to the nature of current training datasets~\cite{hudson2019gqa}, which primarily focus on visual perception rather than spatial understanding. Recent works have attempted to address this challenge through various approaches. For image-based spatial reasoning, several studies have proposed synthetic datasets~\cite{liu-etal-2023-visual,Chen_2024_CVPR,cheng2024spatialrgpt,cai2025spatialbotprecisespatialunderstanding}. For video-based spatial reasoning, approaches have emerged either by incorporating 3D representations as bridging knowledge~\cite{zhu2025llava3dsimpleeffectivepathway} or by tracking objects across frames~\cite{liu2024coarsecorrespondencesboostspatialtemporal}. However, systematic approaches to enhance LVLMs' spatial reasoning capabilities that can generalize across both images and videos remain largely unexplored. Our work fills this gap by developing a principled reasoning framework in LVLMs.

\section{Methodology}
In this section, we present our approach to advancing spatial reasoning capabilities in LVLMs. Given a spatial reasoning question $Q$ and a visual input $I=\{I_n\}_{n=1}^N$, where each $I_n$ represents a single image, $N=1$ corresponds to a single image input, and $N>1$ denotes a video sequence or multiple image inputs, our goal is to enable LVLMs to derive the answer $A$ through iterative visual drawing and thinking. We first introduce our reasoning paradigm (\ref{subsec:dris}), and then detail the training framework (\ref{subsec:training}) that cultivates this reasoning capability. 

\subsection{Drawing to reason in space}
\label{subsec:dris}
We propose ``{drawing to reason in space,}'' a reasoning paradigm that empowers LVLMs to decompose complex spatial reasoning tasks into a sequence of interpretable visual drawing and thinking steps, as illustrated in Figure~\ref{fig:paradigm}. Formally, given the visual input $I$ and question $Q$, the LVLM $\mathcal{M}$ generates a multi-step reasoning path $R$ to derive the final answer. The reasoning path $R$ can be represented as a $T$-step chain: $R=\{(r_t,e_t,o_t)\}_{t=1}^T$, where each step interleaves natural language reasoning $r_t$, drawing operations $e_t$, and observed results $o_t$ from executing the drawing operations~\cite{yao2022react}. This iterative process continues until $\mathcal{M}$ reaches a conclusive answer in the final reasoning step $r_T$. Next, we elaborate on the core components of the paradigm.

\paragraph{Drawing operations for spatial reasoning.} We equip LVLMs with two essential drawing operations $\mathcal{T} = \{\mathcal{T}_\text{box}, \mathcal{T}_\text{line}\}$ for bounding box annotation and auxiliary line drawing, respectively. These operations are fundamental to spatial reasoning as they enable explicit representation of object locations and their spatial transitions\textemdash bounding boxes anchor object positions while auxiliary lines visualize spatial trajectories and relationships\footnote{Since this work focuses on inter-object spatial relationships rather than object-specific details, we adopt only basic drawing operations rather than common visual manipulations like zooming or cropping~\cite{cheng2024least,openai2025thinking}.}. 
 % 
 %This design mirrors human cognitive processes in spatial analysis. 
Each operation $\tau \in \mathcal{T}$ accepts three parameters:
\begin{itemize}[leftmargin=10pt]
    \item $k:$ the index of the target image from either the original input $I$ or previous drawing outputs $\{o_i\}$;
    \item $p:$ single or multiple spatial coordinates for annotating bounding boxes or drawing auxiliary lines;
    \item $l:$ semantic labels describing the annotated content.
\end{itemize}
The execution output of each drawing operation is an annotated version of the target image, preserving the original content while overlaying the specified visual elements. While this minimal operation set proves effective for spatial reasoning, it can be readily extended with additional visual manipulation tools in future work.

\paragraph{Per-step spatial reasoning.} At the $t$-th reasoning step, $\mathcal{M}$ generates both natural language reasoning $r_t$ and drawing operations $e_t$ based on the entire interaction history:
\begin{equation}
\big(r_t,e_t=\{e_t^j=(k_t^j,p_t^j,l_t^j)\}_{j=1}^{m_t}\big) = \mathcal{M}(I, Q, R_{<t}),
\end{equation}
where $e_t$ specifies a set of $m_t$ operations to be executed, $e_t^j$ denote each single operation with three necessary parameters, and $R_{<t}=\{(r_i,e_i,o_i)\}_{i<t}$ means the reasoning history before step $t$. The operations in $e_t$ can target different images through distinct indices $k_t^j$, which is crucial for both image and video reasoning as key information may come from multiple previous drawing outputs or different video frames. After execution, these operations output a set of annotated images $o_t=\{o_t^j\}_{j=1}^{m_t}$, which are sequentially indexed to maintain order. To enable index-based image retrieval for drawing, $\mathcal{M}$ has access to the indices of all available images, including both the original visual input and operation-generated outputs. The process terminates when $r_t$ reaches a final answer, in which case $e_t=\emptyset$ and $o_t=\emptyset$.

% \paragraph{Per-Step Decomposition.} 对于每一个reasoning step，模型的输入是xxx，输出是单步的自然语言推理过程和要调用的工具的参数；每一步可以调用多个工具处理多张图片；原始输入和工具产生的新图片的index都会被显示地告诉模型，以便模型之后进行操作。

% , where each step either performs manipulation on $I$ (e.g., marking an area with a bounding box) or

\subsection{Training framework}\label{subsec:training}
\begin{figure}[!t]
  \centering
  \includegraphics[width=\linewidth]{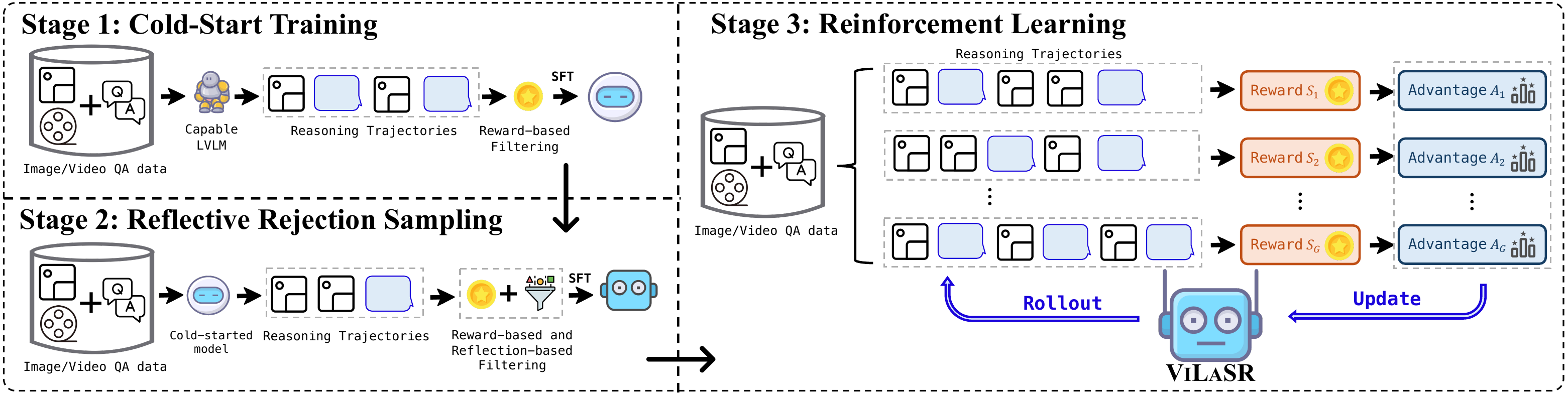}
  \vspace{-15pt}
  \caption{Overview of the three-stage training framework of \textsc{ViLaSR}.}
  \label{fig:ViLaSR_train}
\end{figure}

Our training framework consists of three stages: cold-start training with synthetic offline data to establish basic visual interaction capabilities, reflective rejection sampling to cultivate reflection behaviors, and reinforcement learning for incentivizing reasoning potentials, as illustrated in Figure~\ref{fig:ViLaSR_train}. During training, we focus on multiple-choice questions (e.g., ``Which object is to the left of the chair? A. table ...'') and numerical questions (e.g., ``How many tables appear in this video?'') due to their amenability to automated correctness evaluation, strategically excluding free-form answer questions (e.g., describing motion trajectories). In what follows, we first present our reward function design that guides all three stages, followed by detailed descriptions of the stage-wise training procedures.

\paragraph{Reward function.} 
For reward design, we propose a rule-based function that combines answer correctness and reasoning format adherence:
\begin{equation}
    S =\mathbbm{1}\big(S_\text{correct} > \beta\big)\cdot\big(S_\text{correct}(A,\hat{A}) + S_\text{format}(R)\big),\label{reward_all}
\end{equation}
where $\beta$ is a threshold that ensures format rewards are only granted when a minimum correctness threshold is met, preventing reward hacking where the model might optimize for format adherence at the expense of task accuracy. $\hat{A}$ is the predicted answer that we extract from the model's final reasoning step $r_T$ using rule-based parsing. 
% as they are difficult to automatically evaluate. 
Specifically, we compute $S_\text{correct}$ as follows:
\begin{equation}
    S_\text{correct}({A}, \hat{A}) =\begin{cases}
        \mathbbm{1}(A = \hat{A}), & \text{for multiple-choice questions,} \\
        \frac{1}{|C|} \sum_{\theta \in \mathcal{C}} \mathbbm{1} \left( \frac{|A - \hat{A}|}{A} < 1 - \theta \right), & \text{for numerical questions.}
    \end{cases}\label{reward_correct}
\end{equation}
For multiple-choice questions, we simply check the exact match between the predicted and ground-truth answers. For numerical questions, we adopt Mean Relative Accuracy (MRA)~\citep{yang2024think} for reward computation, which provides a more robust evaluation than conventional metrics like absolute error or fixed thresholding. Instead, MRA examines the prediction accuracy across multiple confidence levels $\mathcal{C} = \{0.50, 0.55, \ldots, 0.95\}$. For each threshold $\theta\in\mathcal{C}$, it checks if the relative error between predicted value $A$ and ground truth $\hat{A}$ falls below $(1-\theta)$. The final score averages these binary outcomes, effectively measuring the model's precision across different stringency levels.

For $S_\text{format}$, we evaluate the quality of reasoning format based on the structural validity of $R$, assigning a score of 1 if all operations in the reasoning path $R$ are executable, and 0 otherwise.
% For $S_\text{format}$, we evaluate both the structural validity and reflective nature of drawing operations:
% \begin{align}
%     S_{\rm format}(A)=\frac{1}{2}\big(\mathbbm{1}(\text{All operations from } R\text{ are executable})+\mathbbm{1}(R\text{~satisfies condition~\ref{condition_reflection}})\big).
% \end{align}
% \gnote{consider putting the reward at the beginning of the section.}

\paragraph{Cold-start training.} Prompting alone often fails to elicit effective visual manipulation abilities in LVLMs for reasoning~\cite{gupta2023visual,cheng2024least}. Therefore, we initialize models' ability to reason in space with drawing operations through supervised learning on a synthetic dataset $\mathcal{D}_\text{cold}$. The training objective is to minimize the average negative log-likelihood over all reasoning and operation tokens:
\begin{equation}
\mathcal{L}_\text{cold} = \mathbb{E}_{(I,Q,A,R=\{(r_t,e_t,o_t)\}_{t=1}^T)\sim\mathcal{D}_\text{cold}} \big(-\frac{1}{N}\sum_{t=1}^T\log p(r_t,e_t|I,Q,R_{<t})\big)
\end{equation}
where $N=\sum_{t=1}^T(|r_t|+|e_t|)$ denotes the total number of tokens in reasoning steps and drawing operations. To construct $\mathcal{D}_\text{cold}$, we first collect a diverse set of image and video question-answering pairs from publicly available datasets, comprising visual inputs $I$, questions $Q$, and ground-truth answers $A$. We then leverage Qwen2.5-72B-VL~\cite{bai2025qwen2} to generate reasoning paths following our reasoning paradigm described in \S\ref{subsec:dris}. The generated paths are subsequently filtered based on rule-based correctness and format checking in Eq.~\ref{reward_all} to ensure high-quality demonstrations of spatial reasoning. Appendix~\ref{appendix:data} shows more details.

% i\leqslant m_{t_1},j\leqslant m_{t_2}
\paragraph{Reflective rejection sampling.} The success of RL often relies on the model's initial capability to exhibit reflective behaviors~\cite{gandhi2025cognitivebehaviorsenableselfimproving}. In our case, the ability to reflect on and revise drawing operations based on observed execution output is crucial. Accordingly, we define reflective behavior as the recurrence of identical labels across different time steps within the same reasoning process. Formally, for a reasoning path $R$, reflection occurs when:
\begin{align}
    \exists (t_1,t_2,u,v): (l_{t_1}^u &= l_{t_2}^v) \land (e^u_{t_1}\not=e^v_{t_2}) \text{ in } R=\{(r_t,e_t=\{e_t^j=(k_t^j,p_t^j,l_t^j)\}_{j=1}^{m_t},o_t)\}_{t=1}^T\label{condition_reflection},
\end{align}
where $l_{t_1}^i$ and $l_{t_2}^j$ represent semantic labels assigned to drawing operations at different reasoning steps, and $e_t^j$ denote the parameters of one drawing operation.  However, after cold-start training, we observe that the resulting model, denoted as $\mathcal{M}_\text{cold}$, infrequently exhibits such reflective behavior~(see \S\ref{ablation} for more details), potentially limiting the effectiveness of subsequent RL optimization. To address this limitation, we introduce a novel reflective rejection sampling mechanism. Given a batch of spatial reasoning examples $\mathcal{D}_\text{reflect}$, we first use $\mathcal{M}_\text{cold}$ to sample corresponding reasoning paths, and then continue fine-tune $\mathcal{M}_{\text{cold}}$ with the following objective:
\begin{align}
\mathcal{L}_\text{reflect} &= \mathbb{E}_{(I,Q,A) \sim \mathcal{D}_\text{reflect}, R \sim \mathcal{M}_{\text{cold}}(\cdot | I, Q)} \big(-\frac{1}{N}\sum_{t=1}^T\log p(r_t,e_t|I,Q,R_{<t})\big)\phi,
\end{align}
where $\phi$ acts as a binary filter that equals 1 if and only if the reasoning path $R$ not only yields the correct answer $A$ and meets format criteria, but also satisfies the reflection condition in Eq.~\ref{condition_reflection}. This selective training strategy encourages the model to learn from high-quality reasoning paths that demonstrate both reflective thinking and correct reasoning, facilitating the development of self-correction capabilities crucial for subsequent RL optimization.
% This pattern indicates the model's ability to revise annotations based on observed execution outputs. 
    % \mathcal{L}_\text{reflect} = \mathbb{E}_{(I,Q,A,R=\{(r_t,e_t,o_t)\}_{t=1}^T)\sim\mathcal{D}_\text{reflect}} \big(-\frac{1}{N}\sum_{t=1}^T\log p(r_t,e_t|I,Q,\{(r_i,e_i,o_i)\}_{<t})\big).
% \phi&=\mathbbm{1}\big(R\text{~satisfies condition~\ref{condition_reflection} and yields correct answer}A\big),

\paragraph{Reinforcement learning.} 
% We further optimize the model through RL with carefully designed rollout policies and reward functions. During policy rollout, we monitor the reasoning process and enforce early termination to prevent inefficient or circular reasoning patterns. Specifically, we set the reward to zero and terminate when: (1) the model fails to invoke any tools in the current step; (2) the number of accumulated images exceeds a predefined threshold; or (3) a tool operation duplicates any previous one. The reward function we used include two parts: accuracy reward $S_\text{acc}$ and format reward $S_\text{format}$.
We further optimize the model using RL with carefully designed rollout policies. During policy rollout, we monitor the reasoning process and apply early termination to avoid inefficient or circular reasoning patterns. Specifically, the rollout is terminated and the reward is set to zero when any of the following conditions are met: (1) the model fails to generate any drawing operations, i.e., $e_t=\emptyset$ while $r_t$ has not reached a final answer; (2) the number of accumulated images exceeds a predefined threshold $\alpha$; or (3) a drawing operation duplicates a previously executed one, i.e., $\exists t_1,t_2,u,v: e^u_{t_1}=e^v_{t_2}$.

With the rollout policies defined above and reward functions in Eq.~\ref{reward_all}, we optimize the policy using GRPO~\cite{deepseekai2025deepseekr1incentivizingreasoningcapability} without the KL penalty term~\cite{hu2025openreasonerzeroopensourceapproach}:
\begin{align}
    \mathcal{L}_\text{RL} &= \mathbb{E}_{\substack{(I,Q,A) \sim \mathcal{D}_\text{rl} \\ \{R_i\}_{i=1}^G \sim p_{\text{old}}(\cdot | I, Q)}} \left( -\frac{1}{G} \sum_{i=1}^G \frac{1}{N_i} \sum_{t=1}^{T_i} \min \left( \rho_{i,t} A_{i}, \, \text{clip}(\rho_{i,t}, 1-\epsilon, 1+\epsilon) A_{i} \right) \right), \\
    \rho_{i,t} &= \frac{p(r_{i,t}, e_{i,t} | I, Q, R_{i,<t})}{p_{\text{old}}(r_{i,t}, e_{i,t} | I, Q, R_{i,<t})},     A_{i} = \frac{S_i - \text{mean}(\{S_j\}_{j=1}^G)}{\text{std}(\{S_j\}_{j=1}^G)},
    % A_{i,t} &= \frac{S_{i} - \text{mean}(\{S_j\}_{j=1}^G)}{\text{std}(\{S_j\}_{j=1}^G)},
\end{align}
where $G$ is the number of rollout reasoning paths, $R_i=\{(r_{i,t},e_{i,t},o_{i,t})\}_{t=1}^{T_i}$ is the $i$-th reasoning path, $N_i$ denotes the total length of the $R_i$ except the tool outputs, $S_i$ is the reward of $R_i$, and $p$ and $p_\text{old}$ represent the current and old policy distributions, respectively. The normalized score $A_{i,t}$ reflects the relative quality of each reasoning path within the rollout group, enabling the model to distinguish between learnable and poor reasoning trajectories.
\section{Experiment}

\subsection{Experimental setups}\label{experimental_setup}

\paragraph{Evaluation benchmark and metrics.} To evaluate the effectiveness and generalization of \textsc{ViLaSR}, we conduct experiments on five benchmarks covering three categories: 
% image-based spatial reasoning (), video-based spatial reasoning (), and multi-view spatial reasoning ():
 % This diverse set of benchmarks enables comprehensive evaluation across varying complexity levels. Specifically, the benchmarks include:
% To evalauate the effectiveness and generalization of our proposed method, we conduct experiments on five benchmark datasets that cover two categories of tasks: image-based spatial reasoning and video-based spatio-temporal reasoning. The former emphasizes static spatial understanding and reasoning within fixed environments, while the latter introduces temporal dynamics in changing environments, thereby increasing the overall reasoning complexity. 
\begin{itemize}[leftmargin=10pt]
    \item \textbf{Image spatial reasoning}: focusing on static relationships and sequential planning, including (1) Maze~\citep{ivanitskiy2023configurable}, specifically designed for navigation assessment; (2) SpatialEval-Real~\citep{wang2024spatial}, demanding real-world spatial relation understanding.
    % which contains four diverse tasks, %i.e., ``Spatial-Map,'' ``Maze-Nav,'' ``Spatial-Grid,'' and ``Spatial-Real,'' 
    % ranging from relation understanding to navigation and object counting; and (3) EmbSpatial-bench~\citep{du-etal-2024-embspatial}, focusing on six ego-centric spatial relationships (``above,'' ``below,'' etc.).\gnote{update benchmarks} %``left,'' ``right,'' ``close,'' and ``far'').
    \item \textbf{Video spatial reasoning}: requiring temporal relationship tracking, including VSI-Bench~\citep{yang2024think}, which tests visual spatial understanding over temporal sequences. 
    \item \textbf{Multi-view spatial reasoning}: challenging models to integrate information from multiple perspectives, including SPAR-Bench~\cite{zhang2025flatlandspaceteachingvisionlanguage}, MMSI-Bench~\cite{yang2025mmsibenchbenchmarkmultiimagespatial}.
    % and ViewSpatial Bench~\cite{li2025viewspatialbenchevaluatingmultiperspectivespatial}.\gnote{update}
    
    % integrating information across multiple different images,  which evaluates spatial understanding across different viewpoints of the same scene, 
\end{itemize}
    % evaluates visual spatial reasoning over videos. .
    % which contains four tasks (Spatial-Map, Maze-Nav, Spatial-Grid, and Spatial-Real), involving diverse aspects of spatial reasoning, including relationship, navigation, position understanding, and object counting. EmbSpatial-bench~\citep{du-etal-2024-embspatial} focus on six spatial relationships described from the ego- centric perspective, including above, below, left, right, close and far. And a synthetic maze dataset~\citep{ivanitskiy2023configurable} is constructed specifically for navigation assessment.
These benchmarks consist exclusively of multiple-choice and numerical questions. We evaluate the model performance using accuracy for multiple-choice questions and Mean Relative Accuracy (MRA)~\citep{yang2024think} for numerical questions. Notably, these metrics follow the same formulation as our training objective (Eq.~\ref{reward_correct}). Appendix~\ref{benchmark_stat} shows the benchmark statistics.
% All benchmarks are evaluated using either a multiple-choice answer (MCA) format or a numerical answer (NA) format.
% For MCA tasks, we report accuracy by comparing the model's predicted answer with the ground truth.
% For NA tasks, we adopt the \textit{Mean Relative Accuracy (MRA)} metric, following the definition in~\citep{yang2024think}.

\paragraph{Implementation details. } 
% We adopt Qwen2.5-VL-7B-Instruct~\citep{bai2025qwen2} as the backbone of our framework.  To balance training efficiency and performance, each image and frame is processed at a maximum resolution of $448\times 448$. For video inputs, we uniformly sample 16 frames per clip. All training phases are conducted on 16 $\times$ NVIDIA A100 (80G) GPUs. During both the initial training phase and the reflective fine-tuning phase, we set the learning rate of the LLM components to $1 \times 10^{-5}$. Both phases are trained for three epochs, taking approximately 24 and 3 hours, respectively.
% Our RL phase leverages the VERL \citep{sheng2024hybridflow} framework, with the training batch size set to 32 and rollout batch size set to 8 for each question.

We implement \textsc{ViLaSR} based on Qwen2.5-VL-7B~\citep{bai2025qwen2}. 
% For computational efficiency while maintaining performance, we process all visual inputs at a maximum resolution of $448\times 448$, with video clips uniformly sampled at 16 frames.
During the training phase, we process all visual inputs at a maximum resolution of $256\times 28 \times 28 $, with video clips uniformly sampled at 16 frames.
In the evaluation stage, we maintain the 16-frame count and the $256\times 28 \times 28$ frame resolution for VSI-Bench, while increase the resolution to $448 \times 28 \times 28$ for other benchmarks.
The training is conducted on a cluster of 16 NVIDIA A100 (80G) GPUs. For both cold-start training stage and reflective fine-tuning stage, we optimize the model with a learning rate of $1 \times 10^{-5}$ for three epochs, requiring approximately 24 and 3 hours, respectively. The subsequent RL optimization is implemented using the VERL framework~\citep{sheng2024hybridflow}, where we set the training batch size to 32 and generate 8 candidate reasoning paths per question. We set the maximum cumulative image number $\alpha$  to 42. And the reward threshold $\beta$ in Eq.~\ref{reward_all} is set to 0.0.
% \gnote{check hyper-parameters, $\alpha$, $\beta$}
% \gnote{image/video qa data details}
% \gnote{cold-start data implementation}

\begin{figure}[t]
\captionsetup[subfigure]{labelfont=scriptsize}  % 专门设置标签字体大小
    \centering
    \begin{subfigure}{0.24\textwidth}
        \centering
        \includegraphics[width=\textwidth]{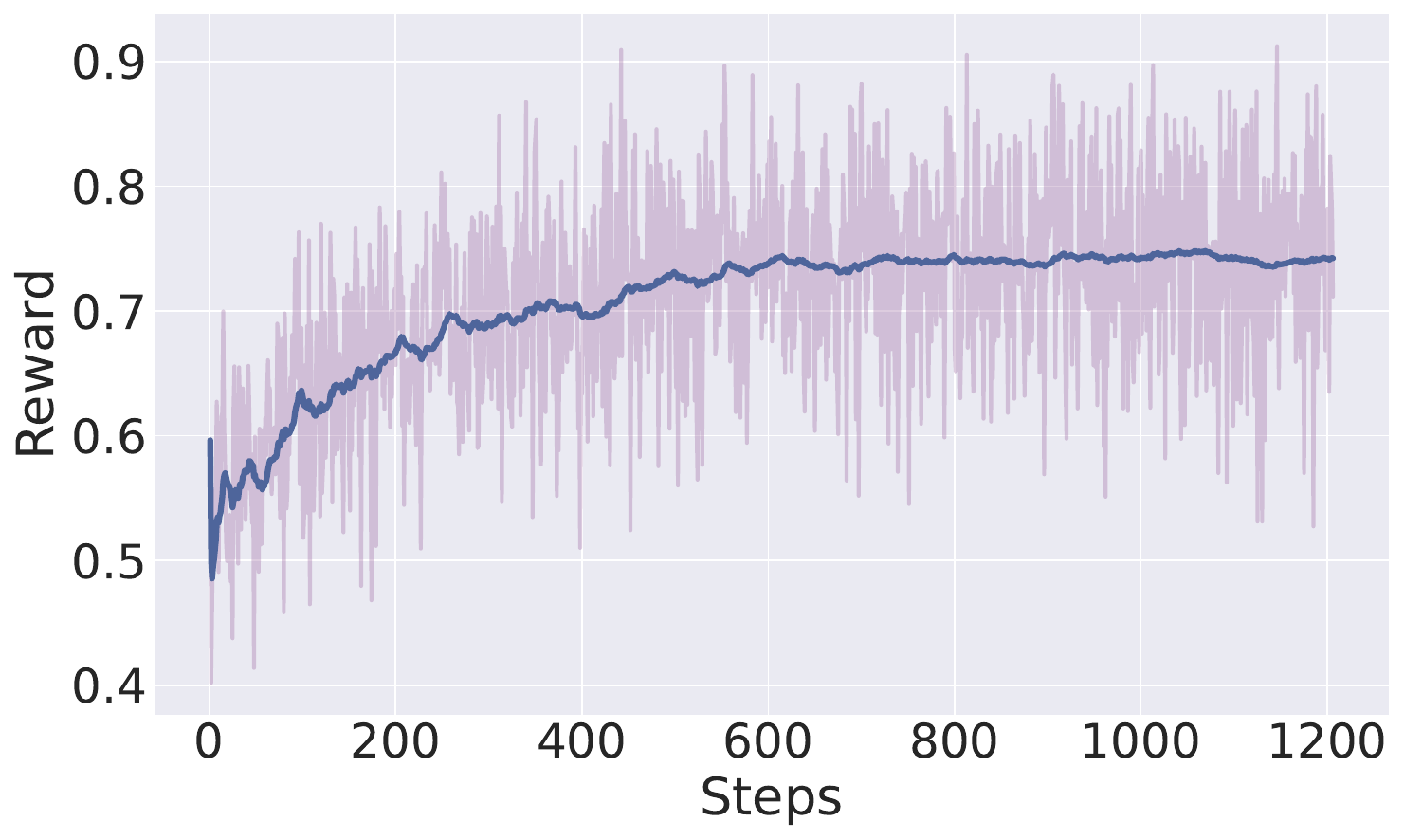}
        \caption{\scriptsize{Overall reward $S$}}
        \label{fig:reward}
    \end{subfigure}
    \hfill
    \begin{subfigure}{0.24\textwidth}
        \centering
        \includegraphics[width=\textwidth]{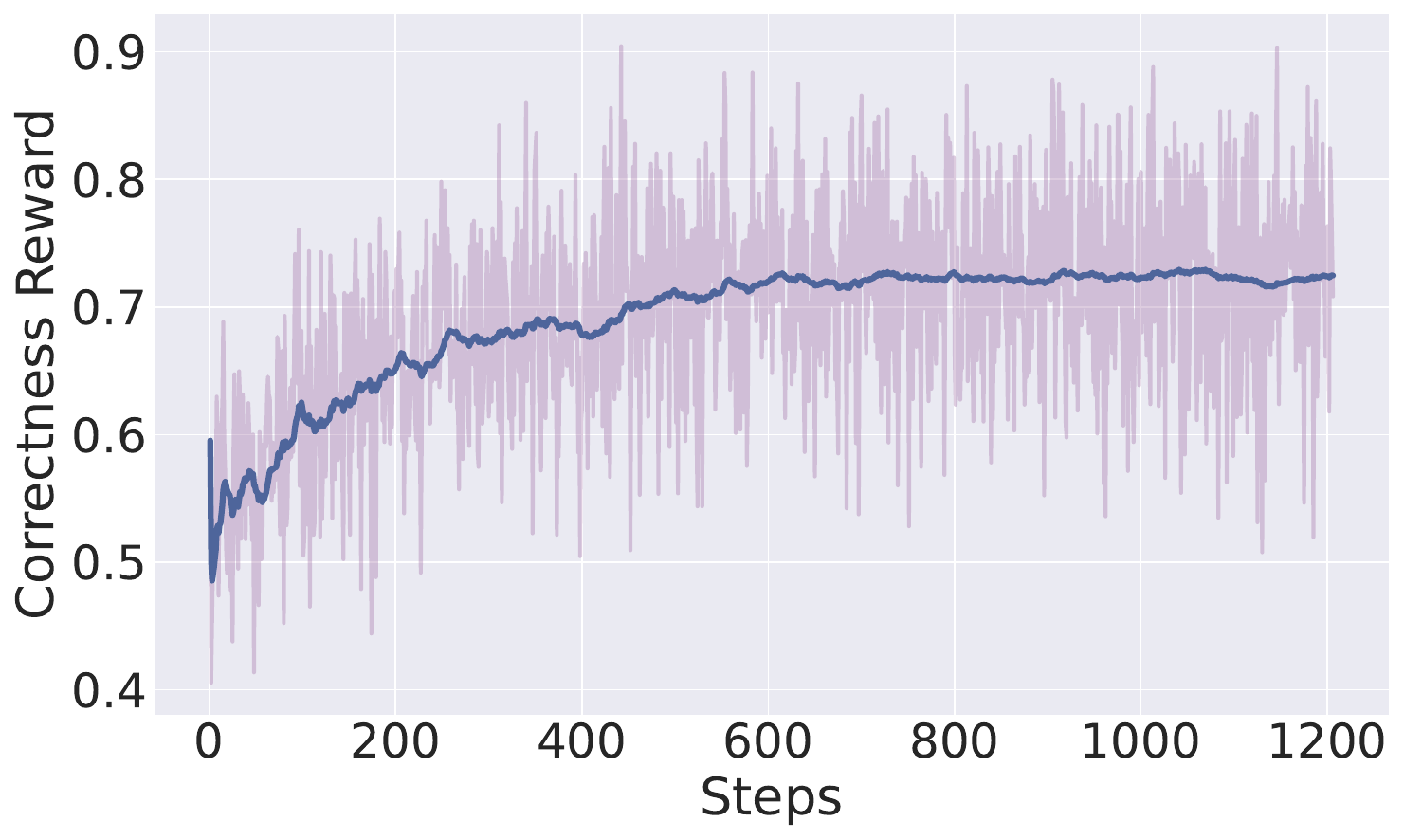}
        \caption{{\scriptsize{Correctness reward $S_\text{correct}$}}}
        \label{fig:acc_reward}
    \end{subfigure}
    \hfill
    \begin{subfigure}{0.24\textwidth}
        \centering
        \includegraphics[width=\textwidth]{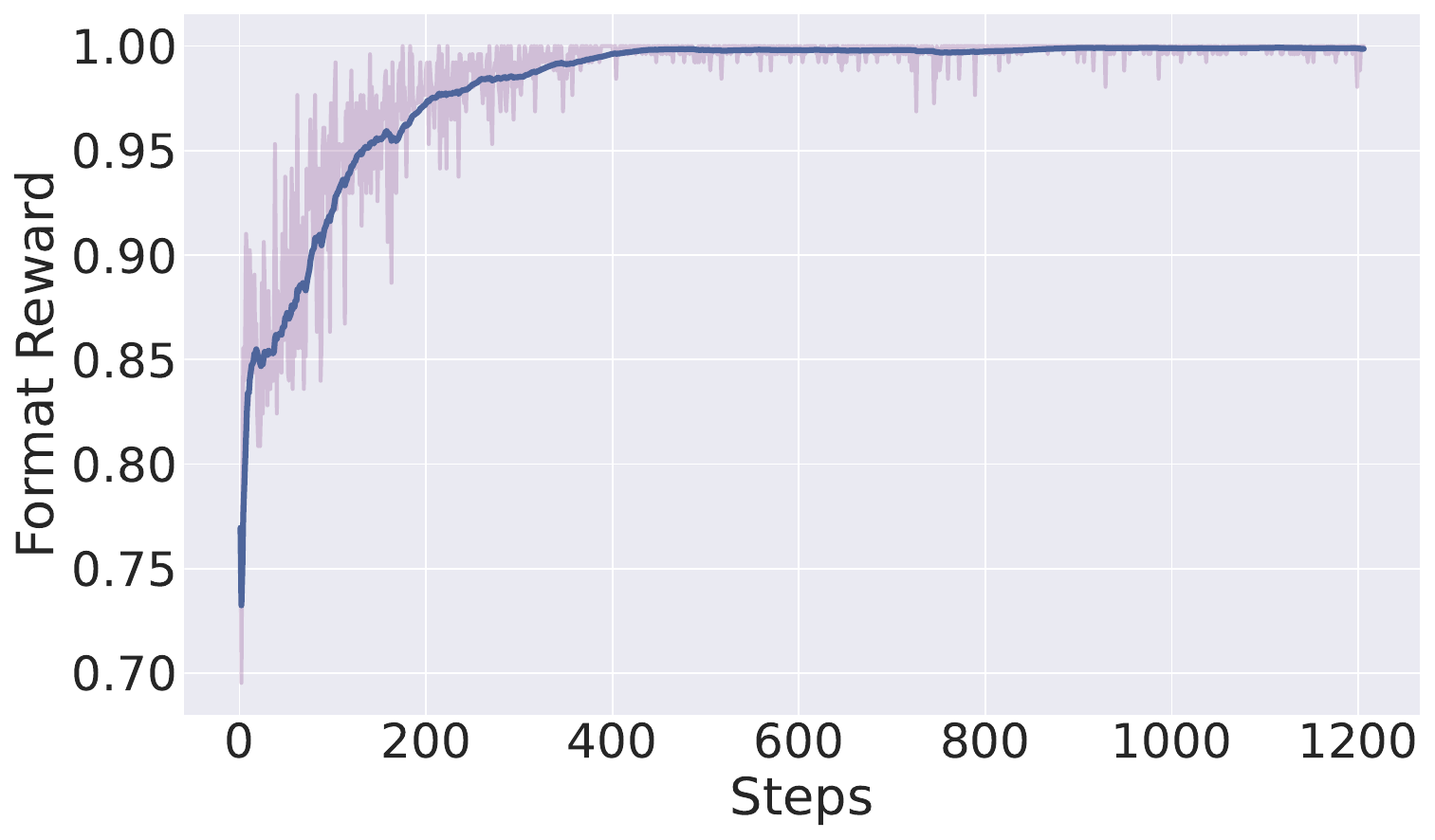}
        \caption{\scriptsize{Format reward $S_\text{format}$}}
        \label{fig:format_reward}
    \end{subfigure}
    \hfill
    \begin{subfigure}{0.24\textwidth}
        \centering
        \includegraphics[width=\textwidth]{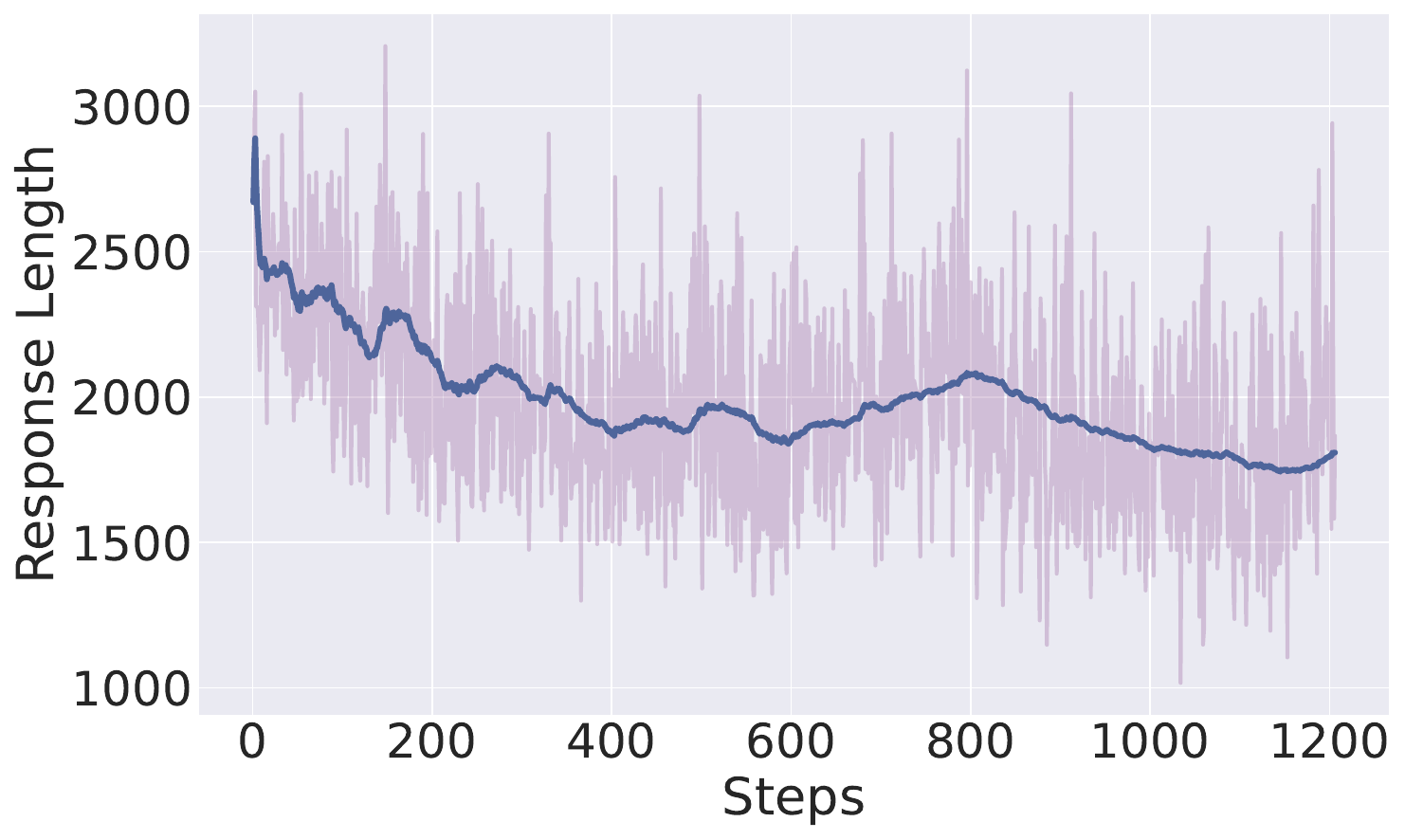}
        \caption{\scriptsize{Response length $N$}}
        \label{fig:response_length}
    \end{subfigure}
    \caption{RL training curves of \textsc{ViLaSR}.}
    \label{fig:curves}
\end{figure}

The training dynamics are shown in Figure~\ref{fig:curves}. As illustrated in (a-c), both overall reward $S$ and its components ($S_\text{correct}$ and $S_\text{format}$) demonstrate steady improvements, validating the effectiveness of our RL optimization. Notably, while achieving better task performance, we observe a decrease in response length $N$ from ~2,500 to 1,800 tokens (d). This can be attributed to two factors: First, during cold-start training, we set a minimum of three reasoning steps in synthetic data to encourage in-depth reasoning, while RL optimization subsequently drives the model toward more efficient use of drawing operations. Second, the limited training data may not fully expose the model to complex scenarios requiring lengthy reasoning chains. We expect that increasing training data diversity could potentially lead to longer but necessary reasoning steps for more challenging spatial tasks~\cite{xie2025logicrlunleashingllmreasoning}.

\paragraph{Baselines. } 

\begin{table}[t]
    \centering
    \caption{Performance comparison across spatial reasoning benchmarks. Gray-shaded rows represent large-sized models ($>$7B parameters). For non-shaded rows, \textbf{bold} and \underline{underlined} numbers indicate the best and second-best results. \textcolor{gray}{\textit{Italic}} numbers in gray-shaded rows indicate performance below non-shaded best results. \textcolor{red}{\textit{Improvement}} refers to the absolute improvement of \textsc{ViLaSR} compared with Qwen2.5-VL-7B w/o reasoning. %Stages 1-3 refer to cold-start training, reflective rejection sampling, and reinforcement learning. 
    \textsuperscript{$\dagger$} and \textsuperscript{$\ddagger$} indicate results from VSI-Bench and VSI-Bench (tiny) set~\citep{yang2024think} respectively. \textsuperscript{$\star$} Results from~\citep{kimiteam2025kimivltechnicalreport}. \textsuperscript{$\star\star$} Results from~\citep{yang2025mmsibenchbenchmarkmultiimagespatial}.  N/A: not support multiple-image input. OpenAI o4-mini is evaluated on a small subset of benchmarks due to its high cost.}%\gnote{check the best results.}}
    % Performance comparison across various spatial reasoning benchmarks. Gray-shaded rows represent either proprietary models or large-sized models ($>$7B parameters). For non-shaded rows, \textbf{bold} and \underline{underlined} numbers indicate the best and second-best results, respectively. \textcolor{gray}{\textit{Italic}} numbers in gray-shaded rows indicate performance below the best result in non-shaded rows. In our ablation studies, ``Stage 1, 2, and 3'' refer to ``cold-start training'', ``reflective rejection sampling'', and ``reinforcement learning'', respectively. \textsuperscript{$\dagger$} indicates results cited from~\citep{yang2024think}. ``N/A'' means the method does not support video inputs.}
    \label{tab:main_tab}
    \resizebox{\linewidth}{!}{
    \begin{tabular}{l cc c c m{0.001em} c m{0.001em} cc}
        \toprule
        \multirow{2}{*}{\textbf{Method}}&\multirow{2}{*}{\textbf{Tool}}&\multirow{2}{*}{\textbf{Reasoning}}&\multicolumn{2}{c}{\textbf{Image}}&&\multicolumn{1}{c}{\textbf{Video}}&&\multicolumn{2}{c}{\textbf{Mutli-view}}\\
        \cmidrule{4-5}
        \cmidrule{7-7}
        \cmidrule{9-10}
         &&& \textbf{MAZE} & \textbf{SpatialEval-Real} && \textbf{VSI-Bench} && \textbf{SPAR-Bench}&\textbf{MMSI-Bench} \\
        \midrule
        \midrule
        \multicolumn{10}{c}{\textbf{\textit{Proprietary LVLMs}}} \\
        \midrule
        % GPT-4o-mini  & - & - & - &  & - \\
        % \rowcolor{gray!20}GPT-4o &\ding{55}&\faCheckSquare& - & - & - && \textcolor{gray}{\textit{35.8}}\textsuperscript{$\dagger$} && - \\
        % \rowcolor{gray!20}Gemini-1.5-Flash &\ding{55}&\faCheckSquare& - & - & - && \textcolor{gray}{\textit{42.1}}\textsuperscript{$\dagger$} && - \\
        \rowcolor{gray!20}GPT-4o &\ding{55}&\ding{55}& \textcolor{gray}{\textit{48.8}} & \textcolor{gray}{\textit{60.7}} && \textcolor{gray}{\textit{34.0}}\textsuperscript{$\dagger$} && \textcolor{gray}{\textit{33.6}} & 30.3$\textsuperscript{$\star\star$}$ \\
        \rowcolor{gray!20}GPT-4o &\ding{55}&\faCheckSquare& \textcolor{gray}{\textit{63.1}} & 65.1 && - && 38.1 & -  \\
        % \rowcolor{gray!20}Gemini-1.5-Flash &\ding{55}&\ding{55}& \textcolor{gray}{\textit{32.4}} & \textcolor{gray}{\textit{59.2}} & 66.4 && \textcolor{gray}{\textit{42.1}}\textsuperscript{$\dagger$} && \textcolor{gray}{\textit{34.9}} \\
        % \rowcolor{gray!20}Gemini-1.5-Flash &\ding{55}&\faCheckSquare& \textcolor{gray}{\textit{58.0}} & \textcolor{gray}{\textit{61.5}} & 67.9 && - && 35.6 \\
        \rowcolor{gray!20}Gemini-1.5-Pro &\ding{55}&\ding{55}& - & -  && \textcolor{gray}{\textit{45.4}}\textsuperscript{$\dagger$} && - & - \\
        \rowcolor{gray!20}Gemini-2.0-Flash &\ding{55}&\ding{55}& \textcolor{gray}{\textit{40.2}} & -  && \textcolor{gray}{\textit{45.4}}\textsuperscript{$\ddagger$} && \textcolor{gray}{\textit{33.4}} & - \\
        \rowcolor{gray!20}Gemini-2.0-Flash &\ding{55}&\faCheckSquare& \textcolor{gray}{\textit{61.7}} & - && - && \textcolor{gray}{\textit{28.0}} & - \\
        \rowcolor{gray!20}OpenAI o3 &\ding{55}&\faCheckSquare& - &- && - && - &  41.0$\textsuperscript{$\star\star$}$ \\
        \rowcolor{gray!20}OpenAI o4-mini &\ding{55}&\faCheckSquare& \textcolor{gray}{\textit{79.0}} & - && - && 46.2 & - \\
        
        \midrule
        \midrule
        \multicolumn{10}{c}{\textbf{\textit{Open-source LVLMs}}} \\
        \midrule
        Qwen2.5-VL-7B &\ding{55}&\ding{55}& 33.7  & 58.5 && 32.7~~ && 31.7 & 26.9~~~~ \\
        Qwen2.5-VL-7B &\ding{55}&\faCheckSquare  & 36.5 & 54.1 && 26.2~~  && 31.6 & 27.1~~~~ \\
        LLaVA-NeXT-Video-7B &\ding{55}&\ding{55}& 34.7 & \textbf{68.1}  && 35.6\textsuperscript{$\dagger$} && 31.3 & 26.8~~~~  \\
        LLaVA-OneVision-7B &\ding{55}&\ding{55}& 30.8 & 62.9  && 32.4\textsuperscript{$\dagger$} && 30.6 & 24.5** \\
        \rowcolor{gray!20}Kimi-VL-A3B-Instruct (16B) &\ding{55}&\ding{55}& 45.0  & {68.9}   && \textcolor{gray}{\textit{37.4}}\textsuperscript{$\star$} && \textcolor{gray}{\textit{33.0}} & - \\
        % \rowcolor{gray!20}InternVL2-40B &\ding{55}&\faCheckSquare& - & - & - && 36.0 & - \\
        \rowcolor{gray!20}Qwen2.5-VL-72B &\ding{55}&\ding{55}& \textcolor{gray}{\textit{50.5}} & \textcolor{gray}{\textit{68.1}}  && \textcolor{gray}{\textit{36.0~~}} && \textcolor{gray}{\textit{37.4}}  & 30.7$\textsuperscript{$\star\star$}$ \\
        % QvQ-72B & - & - & - & - & - \\
        \rowcolor{gray!20}LLaVA-NeXT-Video-72B &\ding{55}&\ding{55}& \textcolor{gray}{\textit{43.2}} & 72.5  && \textcolor{gray}{\textit{40.9}}\textsuperscript{$\dagger$}&& \textcolor{gray}{\textit{35.6}} &  \textcolor{gray}{\textit{28.3}}~~~~ \\
        \rowcolor{gray!20}LLaVA-OneVision-72B &\ding{55}&\ding{55}& \textcolor{gray}{\textit{46.3}} & {{68.8}} && \textcolor{gray}{\textit{40.2}}\textsuperscript{$\dagger$} && \textcolor{gray}{\textit{34.4}} & \textcolor{gray}{\textit{28.4}}$\textsuperscript{$\star\star$}$ \\
        \midrule
        \midrule
        \multicolumn{10}{c}{\textit{\textbf{Representative methods for multimodal reasoning}}}  \\
        \midrule
        \rowcolor{gray!20}\textsc{CogCoM}-17B~\cite{qi2025cogcom} &\faCheckSquare&\faCheckSquare& \textcolor{gray}{\textit{29.8}} & \textcolor{gray}{\textit{49.3}}  && \textcolor{gray}{\textit{N/A}} & &\textcolor{gray}{\textit{N/A}} & \textcolor{gray}{\textit{N/A}}  \\
        VisCoT-7B~\cite{shao2024visual} &\faCheckSquare&\faCheckSquare& 26.0 & 43.2 && N/A && N/A\ & N/A \\
        SpaceR-7B~\cite{ouyang2025spacerreinforcingmllmsvideo} &\ding{55}&\faCheckSquare& \underline{38.6} & 62.7  && \underline{43.5} && \underline{37.1} & \underline{28.8} \\
        \midrule
        \midrule
        \multicolumn{10}{c}{\textbf{\textit{Ours}}} \\
        \midrule
        % \textsc{ViLaSR}&\faCheckSquare&\faCheckSquare& \textbf{89.0} & \underline{63.4} & {\underline{64.9}} && \textbf{42.9} && {\underline{34.9}}\\
         \textsc{ViLaSR}&\faCheckSquare&\faCheckSquare& \textbf{98.2} & \underline{63.9}  && \textbf{45.4} && {\textbf{37.6}} & {\textbf{30.2}}\\
         \textcolor{red}{\textit{Improvement}}&\textcolor{red}{\textit{N/A}}&\textcolor{red}{\textit{N/A}}&\textcolor{red}{\textit{+64.5}}&\textcolor{red}{\textit{+5.4}}&&\textcolor{red}{\textit{+12.7}}&&\textcolor{red}{\textit{+5.9}} & \textcolor{red}{\textit{+3.3}}\\
        % \midrule
        % \textsc{ViLaSR} w/o Stage 3&\faCheckSquare&\faCheckSquare& 85.6 & 63.1 & 64.5 && 38.9 && 34.3 \\
        % \textsc{ViLaSR} w/o Stage 2&\faCheckSquare&\faCheckSquare& \underline{87.1} & 63.9 & \textbf{65.7} && 39.8 && \underline{35.1} \\
        % \textsc{ViLaSR} w/o Stage 2\&3&\faCheckSquare&\faCheckSquare& 85.4  & \textbf{64.1} & 64.3 && 37.1 && 34.6 \\
        % \textsc{ViLaSR} w/o Stage 1\&2\&3   &\faCheckSquare&\faCheckSquare& 28.1 & 53.5 & 55.8 && 17.7 && 23.3 \\        
        % Cold-Start   & 85.9  & 64.1 & 62.5 && 37.1 & - \\
        % Rejection Sampling & - & 62.1 & 64.1 && 41.1 & - \\
        % RL & - & - & - && 42.9 & - \\
        \bottomrule
    \end{tabular}
    }
\end{table}

We compare \textsc{ViLaSR} with various representative models and methods as follows: %\gnote{update baselines}: 
\begin{itemize}[leftmargin=10pt]
    \item \textbf{Proprietary LVLMs}: GPT-4o~\cite{openaigpt4o}, Gemini-1.5-Pro~\cite{team2024gemini}, Gemini-2.0-Flash~\cite{google2024gemini2}, OpenAI o3 and o4-mini~\cite{openai2025o3o4}; 
    \item \textbf{Open-source LVLMs}: These models range from small-sized models including Qwen2.5-VL-7B~\cite{bai2025qwen2}, LLaVA-NeXT-Video-7B~\citep{zhang2024video} and LLaVA-OneVision-7B~\cite{li2024llavaonevisioneasyvisualtask} to large-sized models including Kimi-VL-A3B-Instruct (16B)~\citep{kimiteam2025kimivltechnicalreport}, Qwen2.5-VL-72B~\citep{bai2025qwen2}, LLaVA-NeXT-Video-72B~\citep{zhang2024video}, LLaVA-OneVision-72B~\citep{li2024llavaonevisioneasyvisualtask}; % QvQ-72B~\citep{}
    \item \textbf{Representative models focused on multimodal reasoning}: \textsc{CogCoM}~\citep{qi2025cogcom}, which enables step-by-step visual reasoning through image manipulations with specialized perception tools (e.g., grounding) but is limited to single-image inputs; VisCoT~\citep{shao2024visual}, which first identifies key regions through bounding box annotations followed by textual reasoning chains on single images; and SpaceR~\citep{ouyang2025spacerreinforcingmllmsvideo}, which is a contemporary work that extends spatial reasoning to video understanding through task-specific optimization on automatically curated video QA data.
    % \gnote{SpaceR?}
\end{itemize}

Furthermore, we perform ablation studies to evaluate each individual training stage to assess their contributions to the final performance. Implementation details of the baselines are show in Appendix~\ref{baseline_implementation}.

% (1) Direct Prompting (Direct), where answers are generated without intermediate reasoning steps; (2) Chain-of-Thought Prompting (CoT), which applies the same reasoning prompt as our model but without training; 
\subsection{Main results}
\label{exp:main}

As shown in Table~\ref{tab:main_tab}, \textsc{ViLaSR} shows strong performance across various spatial reasoning benchmarks, from maze navigation to complex video understanding. Our analysis reveals several key findings:
% \gnote{if we have more results on MAZE, we can emphasize the importance of ``drawing to reason in space''}

% \textbf{(1) Superiority of reasoning in visual space.} The experimental results demonstrate a clear advantage of visual-space reasoning over pure textual reasoning. Specifically, OpenAI o3-mini, which incorporates visual manipulation capabilities, exhibits significantly stronger performance compared to GPT-4o and Gemini-1.5-Flash, which process spatial relationships purely in the language domain. This performance differential validates our core hypothesis that sophisticated spatial reasoning necessitates explicit visual operations and intermediate visual state tracking, rather than relying solely on implicit textual representations of spatial relationships.\gnote{if o4-mini is used without tools, we need to delete this paragraph, right?}\textcolor{red}{Ans: Yes, we should delete this paragraph}

\textbf{(1) Limited visual reasoning capabilities in open-source LLMs.} Our experiments reveal a significant disparity between proprietary and open-source LLMs in their ability to leverage reasoning processes. Proprietary models consistently outperform their non-reasoning counterparts by a substantial margin across different benchmarks~(e.g., GPT-4o, Gemini-2.0-Flash). However, open-source models like Qwen2.5-VL-7B show minimal or even negative improvements with reasoning enabled, suggesting a significant gap in their fundamental visual reasoning capabilities. This observation highlights a critical direction for future research: enhancing the multimodal reasoning abilities of open-source models, potentially through improved architectural designs and more sophisticated training strategies that better integrate visual and linguistic information.

\textbf{(2) Strong performance of \textsc{ViLaSR} on image-based spatial tasks.} \textsc{ViLaSR} demonstrates exceptional capabilities in both sequential spatial planning (MAZE) and static spatial understanding (SpatialEval). The remarkable performance on maze navigation, in particular, highlights the advantage of our ``drawing to reason in space'' paradigm: through iterative drawing operations, the model effectively breaks down multi-step navigation sequences into interpretable visual steps, thereby capturing and tracking spatial state transitions. In contrast, existing methods show limited performance due to various constraints: \textsc{CogCoM} is restricted by the capabilities of its perception tools, and VisCoT lacks the ability to reflect and revise its visual operations since it only performs one-time grounding. 

\textbf{(3) Competitive results of \textsc{ViLaSR} on video and multi-view reasoning.} Video and multi-view reasoning present unique challenges beyond static image understanding, as it requires tracking spatial relationships across multiple viewpoints and temporal transitions. While most existing models struggle with this increased complexity, \textsc{ViLaSR} achieves state-of-the-art performance on all benchmarks, surpassing significantly larger open-source models such as LLaVA-OneVision-72B. This success can be attributed to our flexible visual operations that effectively handle dynamic spatial relationships. In contrast, existing approaches face fundamental limitations: \textsc{CogCoM} and VisCoT lack video processing capabilities entirely, while SpaceR's text-centric reasoning fails to fully exploit visual temporal information.

\begin{figure}[t]
    \centering
    \begin{subfigure}{0.58\textwidth}
        \centering
        \includegraphics[width=\textwidth]{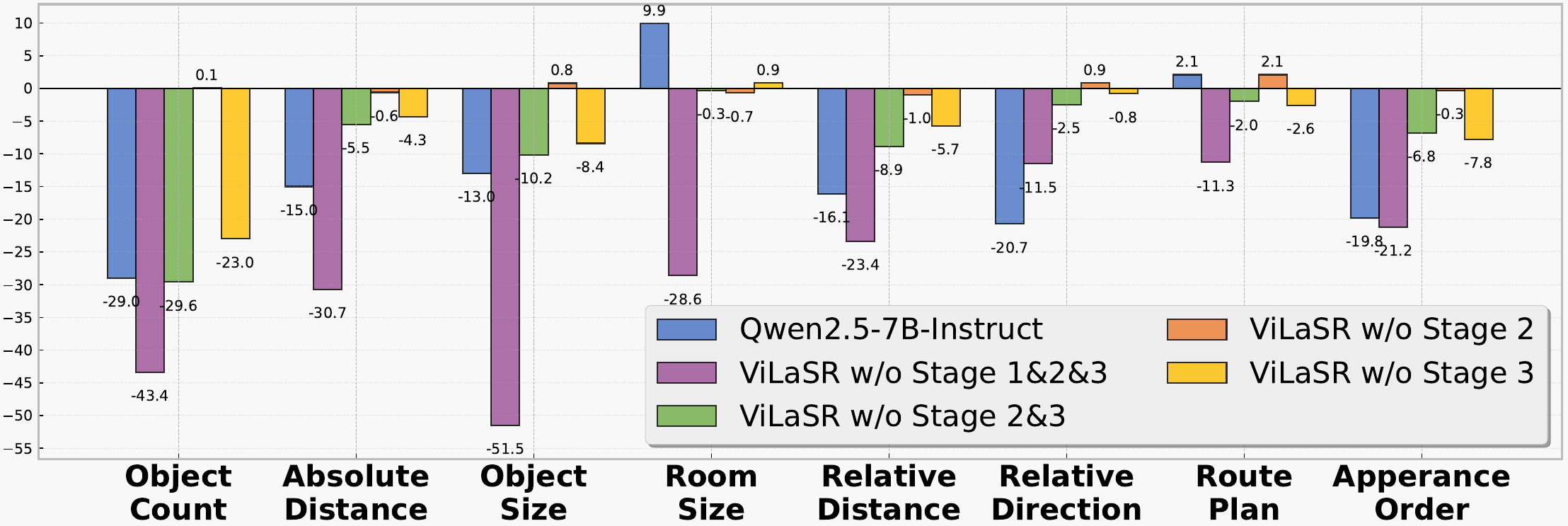}
        \caption{Absolute change in scores compared with \textsc{ViLaSR} on eight subsets of VSI-Bench. }
        \label{fig:ablation}
    \end{subfigure}
    \hfill
    \begin{subfigure}{0.41\textwidth}
        \centering
        \includegraphics[width=\textwidth]{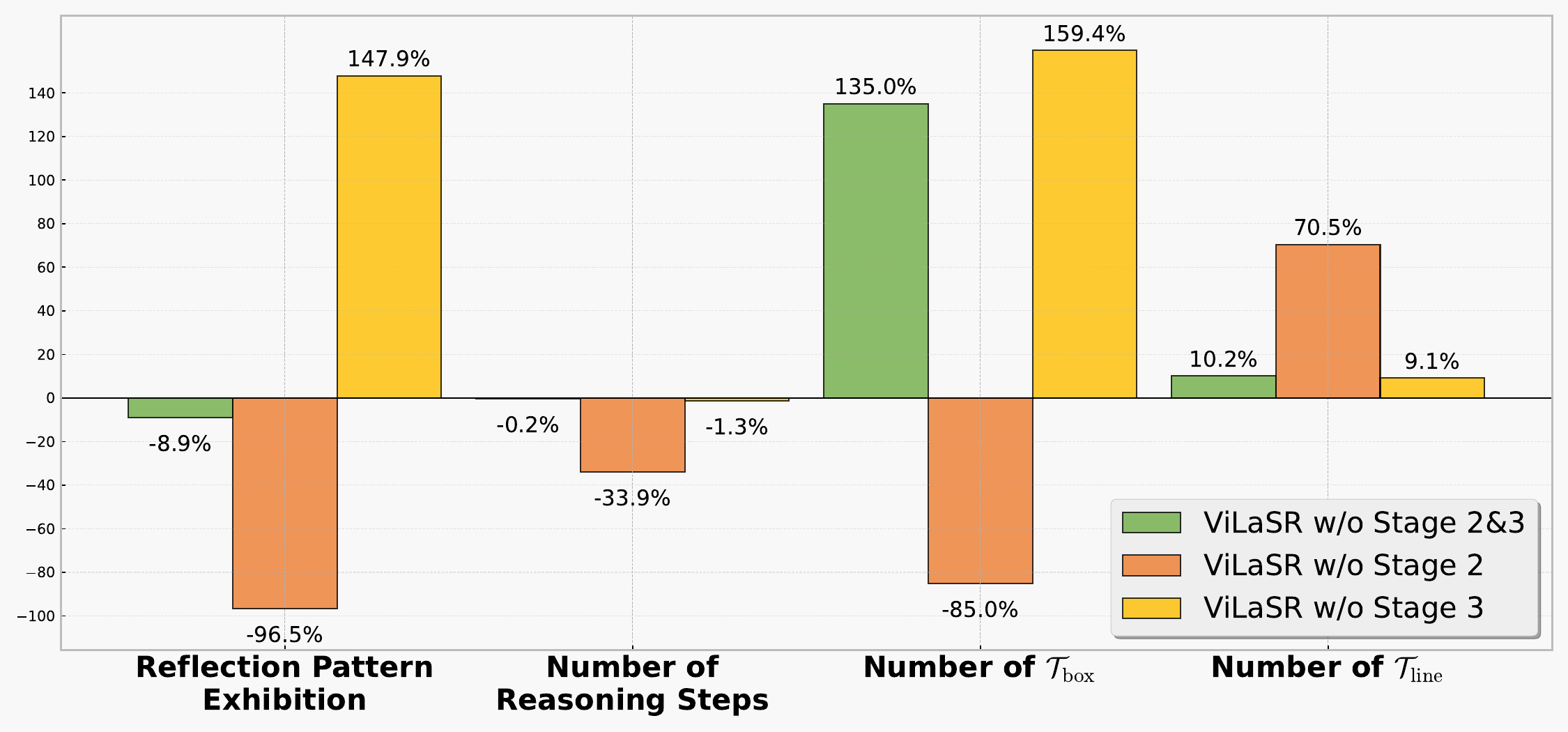}
        \caption{Relative change (\%) in four key behavioral metrics compared with \textsc{ViLaSR}.}
        % \caption{Reflection pattern analysis across different training stages. The bars show the ratio of examples exhibiting reflection after each stage.\gnote{update the results, model name, VSIbench$\to$VSI-Bench}}
        \label{fig:reflection_pattern}
    \end{subfigure}
    % \begin{subfigure}{0.48\textwidth}
    %     \centering
    %     \includegraphics[width=\textwidth]{image/reflection.pdf}
    %     \caption{Reflection pattern analysis across different training stages. The bars show the ratio of examples exhibiting reflection after each stage.\gnote{update the results, model name, VSIbench$\to$VSI-Bench}}
    %     \label{fig:reflection_pattern}
    % \end{subfigure}
    \caption{Ablation study results on VSI-Bench regarding three training stages: cold-start training (Stage 1), reflective rejection sampling (Stage 2), and reinforcement learning (Stage 3).}
    \label{fig:combined}
\end{figure}

\subsection{Ablation study}\label{ablation}
To comprehensively evaluate our training framework, we conduct ablation studies on VSI-Bench's eight spatial reasoning subtasks and analyze four key behavioral metrics. The behavioral metrics include: (1) \textit{Reflection Pattern Exhibition}, measuring the ratio of examples exhibiting reflection behavior as defined in Eq.~\ref{condition_reflection}; (2) \textit{Number of Reasoning Steps}, indicating the average number of reasoning steps ($T$) required to answer each question; and (3,4) \textit{Numbers of} $\mathcal{T}_\text{box}$ \textit{and} $\mathcal{T}_\text{line}$, reflecting the average number of drawing operations utilized per question. The results are shown in Figure~\ref{fig:combined}(a) for task performance and Figure~\ref{fig:combined}(b) for behavioral patterns. We draw the following insights:

\definecolor{modelA}{RGB}{68,114,196}   % #4472C4
\definecolor{modelB}{RGB}{155,78,150}   % #9B4E96
\definecolor{modelC}{RGB}{112,173,71}   % #70AD47
\definecolor{modelD}{RGB}{237,125,49}   % #ED7D31
\definecolor{modelE}{RGB}{255,192,0}    % #FFC000

\textbf{(1) Cold-start training: Essential for spatial reasoning foundation.} Comparing \textcolor{modelB}{``\textsc{ViLaSR} w/o Stage 1\&2\&3''} with \textcolor{modelC}{``\textsc{ViLaSR} w/o Stage 2\&3''}, we observe substantial performance improvements across all subtasks. Notably, when equipped with only our ``drawing to reason in space'' paradigm without training (i.e., \textcolor{modelB}{``\textsc{ViLaSR} w/o Stage 1\&2\&3}''), the model performs even worse than the \textcolor{modelA}{Qwen2.5-VL-7B} backbone. This degradation highlights that sophisticated spatial reasoning capabilities cannot emerge solely with prompting but must be learned through dedicated training.

\textbf{(2) Reflection sampling: Key to self-correction ability.} The comparison between \textsc{ViLaSR} and \textcolor{modelD}{``\textsc{ViLaSR} w/o Stage 2''} reveals the crucial role of reflective rejection sampling. Removing this stage leads to a 96.5\% decrease in reflection pattern exhibition, fundamentally altering the model's behavior and performance. Behaviorally, we observe a 33.9\% reduction in reasoning steps and dramatically different patterns in drawing operation usage: a 85.0\% decrease in $\mathcal{T}_\text{box}$ but a 70.5\% increase in $\mathcal{T}_\text{line}$. Such changes particularly affect tasks requiring precise object localization and measurement: ``Absolute Distance'' (-0.6\%), ``Room Size'' (-0.7\%), and ``Relative Distance'' (-1.0\%). In contrast, tasks that primarily rely on directional judgment or categorical reasoning show minimal impact or even slight improvements (e.g., ``Object Count'': +0.1\%, ``Relative Direction'': +0.9\%), as they can be solved with simpler spatial relationships indicated by lines. This pattern reveals that reflection capability fundamentally shapes how the model builds spatial understanding. Without it, the model shows a tendency to make quick spatial judgments through auxiliary lines without sufficient self-verification. Based on this observation, we hypothesize specific mechanisms through which reflection enhances spatial reasoning: re-examining object locations through bbox annotations and carefully evaluating different spatial relationships before making final judgments (see examples in Appendix~\S\ref{appendix:visualization}). This verification-driven approach, rather than making multiple ``guessing'' attempts through auxiliary lines, appears to be key to accurate spatial reasoning.

\textbf{(3) RL optimization: Dense rewards enable fine-grained learning.} Comparing \textsc{ViLaSR} with \textcolor{modelE}{``\textsc{ViLaSR} w/o Stage 3''}, we observe consistent performance decreases across most subtasks. The substantial increases in both $\mathcal{T}_\text{box}$ (+159.4\%) and $\mathcal{T}_\text{line}$ (+9.1\%) usage without RL suggest that RL optimization helps the model learn to use drawing operations more selectively. Moreover, tasks requiring precise numerical answers (``Object Count,'' ``Absolute Distance,'' ``Object Size'' and ``Room Size'') show greater average performance gaps without RL compared to multiple-choice questions (-9.21\% vs. -4.07\%). This disparity highlights a key advantage of RL optimization: while supervised learning merely maximizes the probability of correct answers, RL provides dense reward signals based on numerical proximity to ground truth, enabling more effective learning for precise spatial measurements.

\subsection{Inference-time scaling}
% Recent studies have demonstrated that inference-time scaling, particularly through multiple parallel sampling and evaluation of the best outcome (commonly known as pass@$k$ evaluation), can effectively boost model performance~\cite{snell2025scaling}. Moreover, research has shown that RL often serves to concentrate the model's pass@$k$ capability into pass@$1$ performance~\cite{yue2025doesreinforcementlearningreally}, suggesting that pass@$k$ metrics can indicate the potential upper bound of model optimization. Motivated by these insights, we analyze the inference-time scaling behavior of \textsc{ViLaSR} and several baseline models to better understand their reasoning capabilities and optimization headroom.

We further evaluate the effectiveness of our training framework through the lens of inference-time scaling analysis such as pass@$k$ evaluation (sampling $k$ outputs in parallel and selecting the best). Prior research has demonstrated: (1) pass@$k$ serves as an empirical upper bound of model capability with sufficiently large $k$~\cite{snell2025scaling}, while pass@$1$ reflects its single-attempt performance; (2) RL has been shown to effectively narrow this gap by consolidating pass@$k$ capabilities into pass@$1$ performance~\cite{yue2025doesreinforcementlearningreally}. These insights motivate us to examine how each training stage contributes to expanding model capability.

% analyze the inference-time scaling behavior of \textsc{ViLaSR} and baseline models to assess the optimization headroom of spatial reasoning capability.

\begin{wrapfigure}[12]{r}{0.5\textwidth}
\vspace{-18pt}
  \centering
  \includegraphics[width=0.5\textwidth]{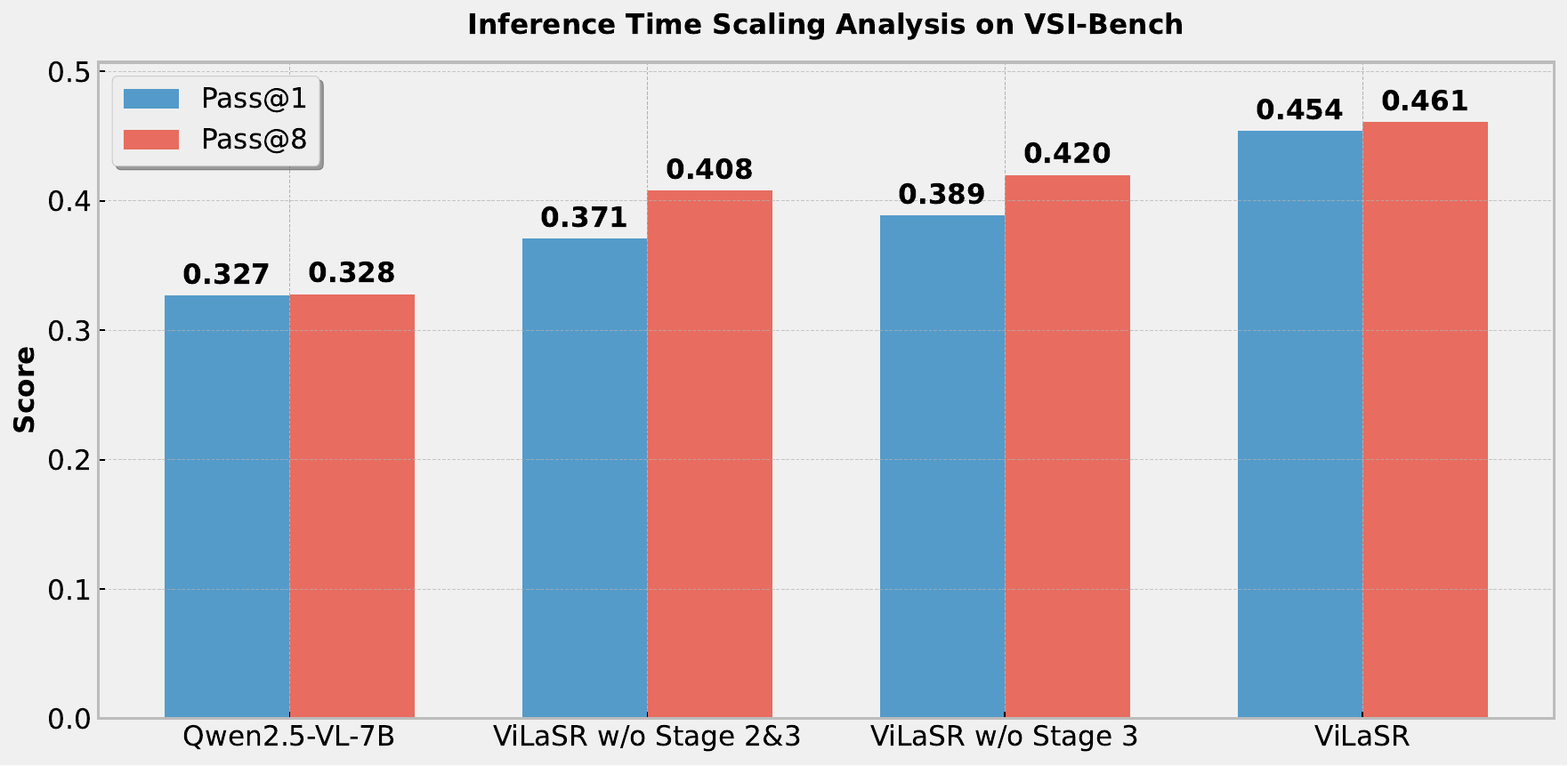}
\vspace{-10pt}
\caption{Analysis of inference-time scaling behavior using pass@$1$ and pass@$8$ performance.}
% We compare pass@1 and pass@8 performance across different model variants.}
  \label{fig:pass_at_k}
% \vspace{-1in}
\end{wrapfigure}

Results in Figure~\ref{fig:pass_at_k} reveal the progressive enhancement of model capability across training stages: (1) Cold-start training substantially improves both the empirical upper bound (pass@$8$) and single-attempt performance (pass@$1$) compared to Qwen2.5-VL-7B; (2) Reflective rejection sampling further elevates both metrics, demonstrating expanded reasoning potential; (3) RL optimization not only pushes the empirical upper bound (pass@$8$) but also significantly narrows the gap with single-attempt performance (pass@$1$), indicating effective consolidation of multi-attempt capabilities that aligns with previous studies about RL's optimization effects~\cite{yue2025doesreinforcementlearningreally}.

\section{Conclusion}

This work presents a principled approach to enhancing spatial reasoning capabilities in LVLMs through visual drawing operations. By enabling direct manipulation in the visual space through the ``drawing to reason in space'' paradigm, we bridge the gap between text-centric reasoning and human-like spatial cognition. Through careful empirical validation, we demonstrate that our three-stage training framework successfully cultivates sophisticated reasoning pattern. Extensive experiments across diverse spatial reasoning benchmarks validate the effectiveness of our approach, showing particular strengths in maze navigation and temporal-spatial understanding tasks. While our results are promising, several important directions remain for future exploration, such as extending our drawing operations to handle more complex 3D spatial relationships, and investigating more efficient training strategies.

\bibliographystyle{plain}
\bibliography{references}
\appendix

% \section{Technical Appendices and Supplementary Material}
% Technical appendices with additional results, figures, graphs and proofs may be submitted with the paper submission before the full submission deadline (see above), or as a separate PDF in the ZIP file below before the supplementary material deadline. There is no page limit for the technical appendices.
\newpage
\section{Appendix outline}

In these supplementary materials, we provide:
\begin{itemize}[leftmargin=10pt]
    \item Dataset construction (Appendix~\ref{appendix:data});
    \item Experimental setup and full evaluation results (Appendix~\ref{appendix:results});
    \item Visualization results (Appendix~\ref{appendix:visualization}).
    \item Complexity comparison between \textsc{ViLaSR} and baselines (Appendix~\ref{complexity});
    \item Discussion on limitations (Appendix~\ref{limitations}) and broader impact (Appendix~\ref{broader_impact}) of \textsc{ViLaSR}.
\end{itemize}
% • Technical details about VSI-Bench construction and
% our linguistic and visual analysis (Appendix B);
% • Evaluation setup and full evaluation results for
% VSI-Bench sub-experiments (Appendix C);
% • Analysis on input sequencing and repetition (Ap-
% pendix D);
% • Additional visualization results (Appendix E).
% \gnote{can we put the concrete results on VSI-Bench (and other benchmarks) in the appendix?}
\section{Dataset construction}
\label{appendix:data}

% 我们构建了用于用于不同阶段训练的数据集，覆盖迷宫、图像、视频空间推理场景。图像数据有利于模型学习在静态空间的绘制辅助线、借助绘图的推理能力，视频数据有助于模型将这些能力扩展到动态场景，捕捉动态和多视角信息，帮助模型获得全面的空间推理能力提升。

% 我们从多个公开数据集中构建或收集数据，并抽样和平衡每个子集的比例。最终构造了ViLaSR-SFT-33k含有推理路径的微调数据集，和ViLaSR-RL-40k用于强化学习训练的数据集。各数据集最终的分布如下：

% ViLaSR-SFT-33k
% 图像：10k（gqa, vsr, openimages, openspaces, spacellava）, 
% maze 7k, 
% video (gpt4scene) 6k,spatial-r1 10k

% ViLaSR-RL-40k
% 图像：10k（gqa, vsr, openimages, openspaces, spacellava）, 
% maze 10k, 
% video (gpt4scene) 10k,spatial-r1 10k.

% 其中，为了获得较高质量的视觉推理路径数据以构造ViLaSR-SFT-33k，我们利用Qwen2.5-VL-72B在原始采样的数据集上生成推理路径，随后过滤出答案正确且格式标准的推理样本，从而作为冷启动数据，这些数据为模型后续RL阶段提高了一个较好的起点。
We created diverse datasets spanning three categories of spatial reasoning tasks: maze navigation that tests path planning abilities, static image understanding that focuses on spatial relationships, and video comprehension that captures temporal evolution and multi-view spatial reasoning. Building upon basic spatial understanding in static images through bounding box annotation and auxiliary line drawing, our video data further extends these capabilities by incorporating temporal dynamics and multi-view perspectives, leading to more comprehensive spatial reasoning abilities.

Our data collection pipeline leverages multiple public datasets, with careful balancing across different task types to ensure comprehensive coverage. This effort resulted in three primary datasets: \textbf{\textsc{ViLaSR}-ColdStart-33k}, a supervised dataset containing curated reasoning paths, \textbf{\textsc{ViLaSR}-RRS-8k}, specifically designed for the reflective rejection sampling stage, and \textbf{\textsc{ViLaSR}-RL-40k}, optimized for reinforcement learning. For \textsc{ViLaSR}-ColdStart-33k construction, we leveraged Qwen2.5-VL-72B~\citep{bai2025qwen2} to generate initial reasoning trajectories following our ``drawing to reason in space'' paradigm detailed in \S\ref{subsec:dris}. A rigorous filtering process based on answer correctness and format validity helped identify the most reliable samples. We applied two format criteria: (1) ensuring all drawing operations are executable and the final answer can be correctly parsed, and (2) requiring at least three rounds of thinking to ensure sufficient spatial reasoning depth. After cold-start training, we constructed \textsc{ViLaSR}-RRS-8k by selecting instances that are generated by the cold-started model and exhibit clear self-correction patterns and correct final answers, providing ideal training examples for cultivating reflective behaviors.

\begin{table}[!ht]
\centering
\caption{Dataset distribution across different training stages.}
\label{tab:dataset-distribution}
\resizebox{0.8\linewidth}{!}{
\begin{tabular}{@{}lccc@{}}
\toprule
\textbf{Subset (Modality)} & \textbf{\textsc{ViLaSR}-ColdStart-33k} & \textbf{\textsc{ViLaSR}-RRS-8k} & \textbf{\textsc{ViLaSR}-RL-40k} \\
\midrule
Maze Navigation (Image) & 7,000 & 1,000 & 10,000 \\
VQA (Image) & 10,000 & 3,000 & 10,000 \\
GPT4Scene~\cite{qi2025gpt4scene} (Video) & 6,000 & 2,000 & 10,000 \\
SR-91k~\cite{ouyang2025spacerreinforcingmllmsvideo} (Video) & 10,000 & 1,800 & 10,000 \\
\midrule
\textbf{Total} & 33,000 & 7,800 & 40,000 \\
\bottomrule
\end{tabular}
}
\end{table}

Table~\ref{tab:dataset-distribution} presents the detailed distribution of these datasets. Our source data encompasses a broad range of spatial reasoning scenarios, drawing from GQA~\cite{hudson2019gqa}, VSR~\cite{zhang2021vsr}, OpenImages~\cite{kuznetsova2020open}, OpenSpaces~\cite{Chen_2024_CVPR} and SpaceLLaVA~\cite{Chen_2024_CVPR}. For maze navigation tasks, we procedurally generated mazes with varying grid sizes (from 3$\times$3 to 6$\times$6) using a depth-first search algorithm\footnote{\url{https://github.com/understanding-search/maze-dataset}}, maintaining approximately equal proportions across different grid sizes. Each maze consists of a clearly marked starting point and four candidate destinations (labeled as A, B, C, and D). The questions are formulated as multiple-choice problems where the model must determine which destination would be reached by following a given action sequence (e.g., ``Determine the final destination from the starting point (green point). Action Sequence: Go up. Go up ...''). This automated generation process ensures diverse maze layouts and action sequences while maintaining problem solvability, creating a controlled environment for evaluating sequential spatial reasoning capabilities.

\section{Experimental setup and full evaluation results}
\label{appendix:results}

\subsection{Baseline implementation}\label{baseline_implementation}
We evaluate \textsc{ViLaSR} against various state-of-the-art models and methods. 

For closed-source LVLMs, we directly query their official APIs as shown in Table~\ref{tab:api}. We evaluate these models using zero-shot prompting by directly providing the input images and questions. Under the setting without reasoning, we explicitly prompt these models to output answers directly without intermediate reasoning steps, i.e., ``Answer with the option's letter from the given choices directly'' for multiple-choice questions and  ``Please answer with a single numerical value (e.g., 42 or 3.14)'' for numerical responses.
\begin{table}[!t]
\caption{APIs used for proprietary models evaluation.}
\label{tab:api}
\centering
\begin{tabular}{lll}
\toprule
\textbf{Model} & \textbf{API Name} & \textbf{Provider} \\
\midrule
GPT-4o & gpt-4o-2024-08-06  & OpenAI \\
Gemini-1.5-Flash & gemini-1.5-flash  & Google \\
Gemini-1.5-Pro & gemini-1.5-pro & Google \\
Gemini-2.0-Flash & gemini-2.0-flash & Google \\
OpenAI o4-mini & o4-mini-2025-04-16 & OpenAI \\
\bottomrule
\end{tabular}
\end{table}

 %For proprietary LVLMs, we directly query the official APIs: ``gpt-4o'' for GPT-4o, ``gemini-1.5-flash'' for Gemini-1.5-Flash, and ``o3-mini'' for OpenAI o3-mini. Similarly, 
For open-source LVLMs ranging from 7B parameters (Qwen2.5-VL-7B~\cite{bai2025qwen2}, LLaVA-NeXT-Video-7B~\citep{zhang2024video}, LLaVA-OneVision-7B~\citep{li2024llavaonevisioneasyvisualtask}) to 72B parameters (Qwen2.5-VL-72B~\citep{bai2025qwen2}, LLaVA-NeXT-Video-72B~\citep{zhang2024video}, LLaVA-OneVision-72B~\citep{li2024llavaonevisioneasyvisualtask}). Kimi-VL-A3B-Instruct(16B)~\citep{kimiteam2025kimivltechnicalreport} leverages Mixture-of-Experts (MoE) architecture with 16.0B total parameters, while dynamically activating 2.8B parameters during inference. We conduct zero-shot evaluation using their standard prompting formats.
% \gnote{please add API name.}

For specialized reasoning models, we evaluate using their officially released checkpoints and prompts: \textsc{CogCoM}~\citep{qi2025cogcom} with its built-in perception tools, VisCoT~\citep{shao2024visual} with its bounding box annotation pipeline, and SpaceR~\citep{ouyang2025spacerreinforcingmllmsvideo} with its video understanding capabilities. 
% Detailed prompting templates and evaluation protocols are provided in Appendix~\ref{baseline_implementation}.

\subsection{Benchmark statistics}\label{benchmark_stat}
Table~\ref{tab:benchmark} shows the benchmark statistics.
\begin{table*}[!t]
% \vspace{-10pt}
    \centering
    \caption{Number of examples in our evaluation benchmark.}%\gnote{update benchmarks}}
    \resizebox{0.8\linewidth}{!}{\begin{tabular}{@{}ccm{0.001em}cm{0.001em}cc@{}}
        \toprule
        \multicolumn{2}{c}{\textbf{Image }}&&\multicolumn{1}{c}{\textbf{Video }}&&\multicolumn{2}{c}{\textbf{Multi-view}}\\
        \cmidrule{1-2}
        \cmidrule{4-4}
        \cmidrule{6-7}
     \textbf{MAZE} & \textbf{SpatialEval-Real}  && \textbf{VSI-Bench} && \textbf{SPAR-Bench} & \textbf{MMSI-Bench} \\
     \midrule
    2,000 & 135 && 5,130&&7,211& 1,000 \\
    \bottomrule
    \end{tabular}}
    \label{tab:benchmark}
    % \vspace{-10pt}
\end{table*}

\begin{table}[t]
    \centering
    \caption{Results on VSI-Bench. Gray-shaded rows represent large-sized models ($>$7B parameters). \textsuperscript{$\dagger$} Results from~\citep{yang2024think}. \textsuperscript{$\ddagger$} Results from VSI-Bench (tiny) set~\citep{yang2024think}. 
    \textsuperscript{$\star$} Results from~\citep{kimiteam2025kimivltechnicalreport}.}
    \label{tab:vsibench}
    \resizebox{\linewidth}{!}{
    \begin{tabular}{l ccccm{0.01em}cccc c}
        \toprule
        \multirow{4}{*}{\textbf{Method}} & \multicolumn{9}{c}{\textbf{Sub-tasks}} & \multirow{4}{*}{\textbf{Average}} \\
        \cmidrule{2-10}
        % & \textbf{Obj. Count} & \textbf{Abs. Dist} & \textbf{Obj. Size} & \textbf{Room Size} & \hlcyan{\textbf{Rel. Dist}} & \hlcyan{\textbf{Rel. Dir}.} & \hlcyan{\textbf{Route Plan}} & \hlcyan{\textbf{Appr. Order}} \\
        &\multicolumn{4}{c}{\textbf{Numerical questions}}&&\multicolumn{4}{c}{\textbf{Multiple-choice questions}}\\
        \cmidrule{2-5}
        \cmidrule{7-10}
        &Object&Absolute&Object&Room&&Relative&Relative&Route&Apperance\\
        &Count&Distance&Size&Size&&Distance&Direction&Plan&Order\\
        \midrule
        \midrule
        \multicolumn{10}{c}{\textbf{\textit{Proprietary LVLMs}}} \\
        \midrule
        \rowcolor{gray!20}GPT-4o & \textcolor{gray}{\textit{46.2}} & \textcolor{gray}{\textit{5.3}} & \textcolor{gray}{\textit{43.8}} & \textcolor{gray}{\textit{38.2}} && \textcolor{gray}{\textit{37.0}} &  \textcolor{gray}{\textit{41.3}} & \textcolor{gray}{\textit{31.5}} & \textcolor{gray}{\textit{28.5}} & \textcolor{gray}{\textit{34.0}}\textsuperscript{$\dagger$} \\
        \rowcolor{gray!20}Gemini-1.5-Flash & 49.8 &  \textcolor{gray}{\textit{30.8}} & 53.5 & 54.4 &&  \textcolor{gray}{\textit{37.7}} & \textcolor{gray}{\textit{41.0}} &  \textcolor{gray}{\textit{31.5}} & \textcolor{gray}{\textit{37.8}} & \textcolor{gray}{\textit{42.1}}\textsuperscript{$\dagger$} \\
        \rowcolor{gray!20}Gemini-1.5-Pro & 56.2 &  \textcolor{gray}{\textit{30.9}} & 64.1 & {{43.6}} && {{51.3}} &  46.3 & {{36.0}} & \textcolor{gray}{\textit{34.6}} & \textcolor{gray}{\textit{45.4}}\textsuperscript{$\dagger$} \\
        \rowcolor{gray!20}Gemini-2.0-Flash & 52.4 & \textcolor{gray}{\textit{30.6}} &  66.7 & \textcolor{gray}{\textit{31.8}} && {{56.0}} &  46.3 & \textcolor{gray}{\textit{24.5}} & 55.1 & \textcolor{gray}{\textit{45.4}}\textsuperscript{$\ddagger$} \\
        \midrule
        \midrule
        \multicolumn{10}{c}{\textbf{\textit{Open-source LVLMs}}} \\
        \midrule
        Qwen2.5-VL-7B & 34.5 & 19.4 & 47.6 & \textbf{40.8} && 32.8 & 24.5 & \underline{32.5} & 29.4 & 32.7~~ \\
        LLaVA-NeXT-Video-7B & 48.5 & 14.0 &  47.8 & 24.2 && \underline{43.5}  & {42.4} &  \textbf{34.0} & 30.6 & 35.6\textsuperscript{$\dagger$} \\
        LLaVA-OneVision-7B & 47.7 & 20.2 & 47.4 & 12.3 &&  42.5 & {35.2} & 29.4  & 24.4 & 32.4\textsuperscript{$\dagger$} \\
        Kimi-VL-A3B-Instruct-16B & - & - & - & - && - & - & - & - & 37.4\textsuperscript{$\star$} \\
        \rowcolor{gray!20}Qwen2.5-VL-72B & \textcolor{gray}{\textit{33.9}} & \textcolor{gray}{\textit{27.2}} & \textcolor{gray}{\textit{59.3}} & \textcolor{gray}{\textit{28.5}} && \textcolor{gray}{\textit{47.2}} & \textcolor{gray}{\textit{35.3}} & \textcolor{gray}{\textit{22.2}} & \textcolor{gray}{\textit{34.5}}& \textcolor{gray}{\textit{36.0}}~~ \\
        \rowcolor{gray!20}LLaVA-NeXT-Video-72B & \textcolor{gray}{\textit{48.9}} & \textcolor{gray}{\textit{22.8}} & \textcolor{gray}{\textit{57.4}} & \textcolor{gray}{\textit{35.3}} && \textcolor{gray}{\textit{42.4}} & \textcolor{gray}{\textit{36.7}} & {{35.0}} & \textcolor{gray}{\textit{48.6}} & \textcolor{gray}{\textit{40.9}}\textsuperscript{$\dagger$} \\
        \rowcolor{gray!20}LLaVA-OneVision-72B & \textcolor{gray}{\textit{43.5}}  & \textcolor{gray}{\textit{23.9}} & \textcolor{gray}{\textit{57.6}} & \textcolor{gray}{\textit{37.5}} && \textcolor{gray}{\textit{42.5}}  & \textcolor{gray}{\textit{39.9}} & \textcolor{gray}{\textit{32.5}} & \textcolor{gray}{\textit{44.6}} & \textcolor{gray}{\textit{40.2}}\textsuperscript{$\dagger$} \\
        \midrule
        \midrule
        \multicolumn{10}{c}{\textbf{\textit{Representative methods for multimodal reasoning}}} \\
        \midrule
        % SpaceR-7B & \underline{60.2} & \underline{28.2} & \underline{59.3} & \underline{31.5} && 40.0 & \textbf{45.5} & 32.0 & \underline{49.0} & \underline{43.2} \\
         SpaceR-7B & \underline{61.9} & \underline{28.6} & \textbf{60.9} & \underline{35.2} && 38.2 & \textbf{46.0} & 31.4 & \underline{45.6} & \underline{43.5} \\
        \midrule
        \midrule
        \multicolumn{10}{c}{\textbf{\textit{Ours}}} \\
        \midrule
        \textsc{ViLaSR} & \textbf{63.5}	& \textbf{34.4}	& \underline{60.6}	& 30.9 && \textbf{48.9}	& \underline{45.2}	& 30.4	& \textbf{49.2} & \textbf{45.4} \\
        % \textsc{ViLaSR} w/o Stage 3 & 43.2 & 44.4 & 27.8 & 41.4 && 40.5 & 30.1 & 52.2 & 31.8 & 38.9 \\
        % \textsc{ViLaSR} w/o Stage 2\&3 & 40.0 & 42.7 & 28.4 & 42.4 && 33.9 & 28.9 & 50.4 & 30.6 & 37.1 \\
        % \textsc{ViLaSR} w/o Stage 1\&2\&3 &  25.5 & 33.7 & 19.1 & 28.0 && 20.1 & 3.7 & 9.1 & 2.3 & 17.7 \\
         \textcolor{red}{\textit{Improvement}}&\textcolor{red}{\textit{+29.0}}&\textcolor{red}{\textit{+15.0}}&\textcolor{red}{\textit{+13.0}}&\textcolor{blue}{\textit{-9.9}}
         &&\textcolor{red}{\textit{+16.1}}&\textcolor{red}{\textit{+20.7}}&\textcolor{blue}{\textit{-2.1}}&\textcolor{red}{\textit{+19.8}}&\textcolor{red}{\textit{+12.7}}\\
        \bottomrule
    \end{tabular}
    }
\end{table}

\subsection{Detailed results on VSI-Bench}
Table~\ref{tab:vsibench} provides detailed results on VSI-Bench. \textsc{ViLaSR} achieves state-of-the-art performance with an average accuracy of 45.4\%, outperforming all baseline methods by a significant margin (+12.7\%). The improvements are particularly pronounced in tasks requiring precise spatial measurements and object localization: ``Object Count'' (+29.0\%), ``Absolute Distance'' (+15.0\%), and ``Object Size'' (+13.0\%). This aligns with our case study observations (see \S\ref{appendix:visualization}) where \textsc{ViLaSR} demonstrates superior capabilities in systematic measurement and spatial reasoning through drawing operations. 

Furthermore, the strong performance in ``Relative Distance'' (+16.1\%) and ``Relative Direction'' (+20.7\%) demonstrates \textsc{ViLaSR}'s effectiveness in comparative spatial reasoning. By explicitly drawing auxiliary lines to connect and measure between objects, our model can more accurately assess relative positions and orientations. The significant improvement in ``Appearance Order'' (+19.8\%) further highlights \textsc{ViLaSR}'s capability in temporal-spatial reasoning, where systematic annotation of objects across multiple frames helps track and verify their sequential relationships.

Interestingly, while showing strong performance in most categories, \textsc{ViLaSR} exhibits slight decreases in ``Room Size'' (-9.9\%) and ``Route Plan'' (-2.1\%). This suggests a limitation in tasks requiring holistic reasoning: ``Room Size'' estimation demands understanding the complete room layout from partial views and reasoning about unseen spaces, while ``Route Plan'' needs global path planning across multiple viewpoints. Unlike localized spatial measurements that can be solved through explicit drawing operations, these tasks require more sophisticated global inference capabilities to integrate information across different views and time points. This reveals potential directions for extending our drawing-based framework: we could introduce specialized drawing tools for layout reconstruction and space completion. Such extensions would be natural additions to our current drawing operation set, enabling more comprehensive spatial reasoning capabilities.

\subsection{Statistical significance analysis}\label{significance}
We conduct rigorous statistical significance tests to validate our experimental results. Using paired $t$-tests, we find that \textsc{ViLaSR} significantly outperforms the base model (Qwen2.5-VL-7B) on MAZE, SpatialEval-Real, VSI-Bench, SPAR-Bench and MMSI-Bench ($p < 0.05$). These results demonstrate the robust and consistent advantages of our approach across different spatial reasoning scenarios.
% NIPS version
% We conduct rigorous statistical significance tests to validate our experimental results. Using paired $t$-tests, we find that \textsc{ViLaSR} significantly outperforms the strongest baseline (Spatial-R1) on MAZE, SpatialEval, EmbSpatial and SPAR-Bench ($p < 0.05$). For VSI-Bench, while \textsc{ViLaSR} achieves marginally better performance, the difference is not statistically significant. These results demonstrate the robust and consistent advantages of our approach across different spatial reasoning scenarios.\gnote{update}

\subsection{Prompt template}

\begin{figure}[!t]
  \centering
  \includegraphics[width=\linewidth]{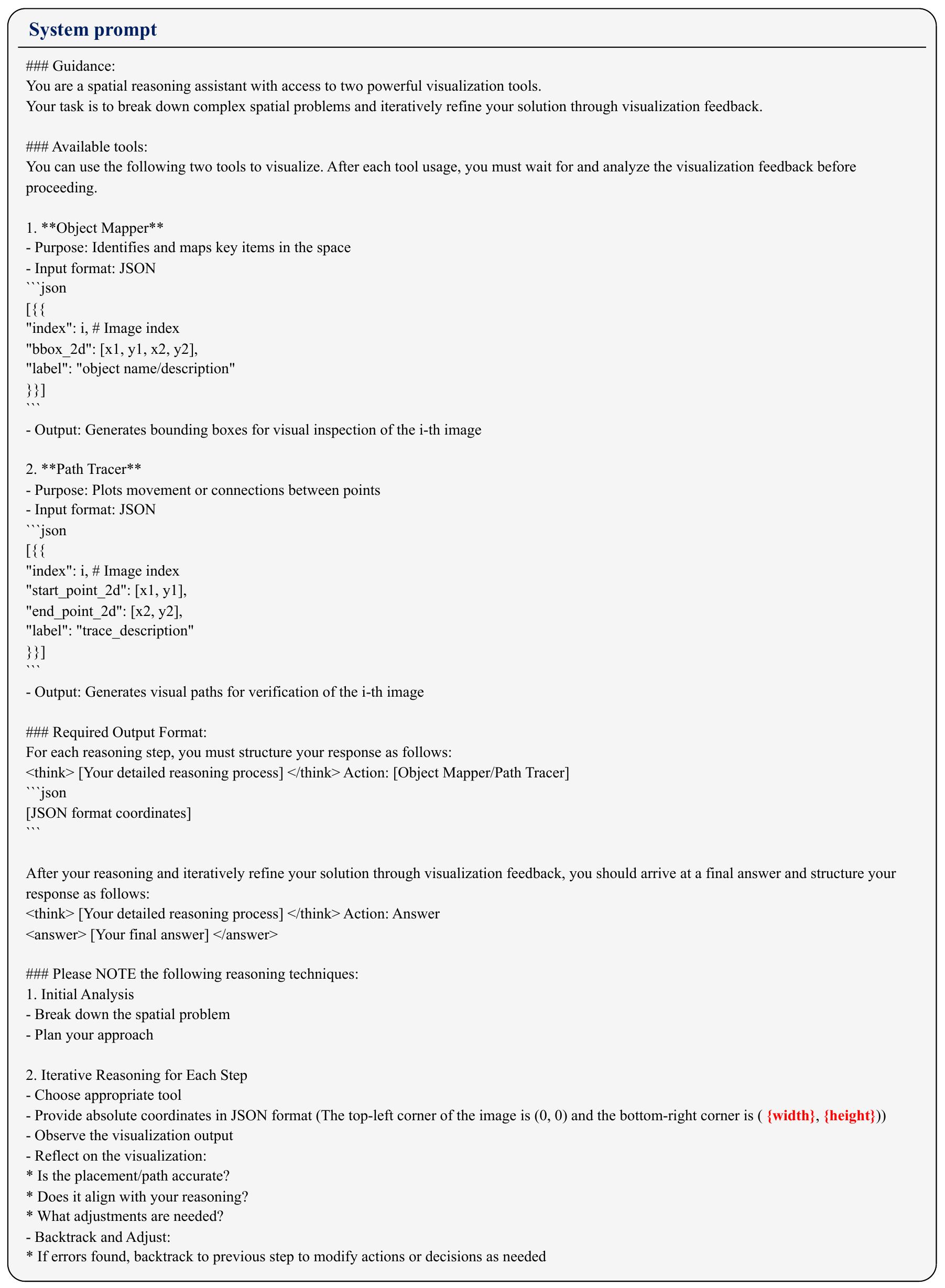}
  \vspace{-20pt}
  \caption{System prompt used in \textsc{ViLaSR}.}
  \label{fig:sys_prompt}
\end{figure}

\begin{figure}[!t]
  \centering
  \includegraphics[width=\linewidth]{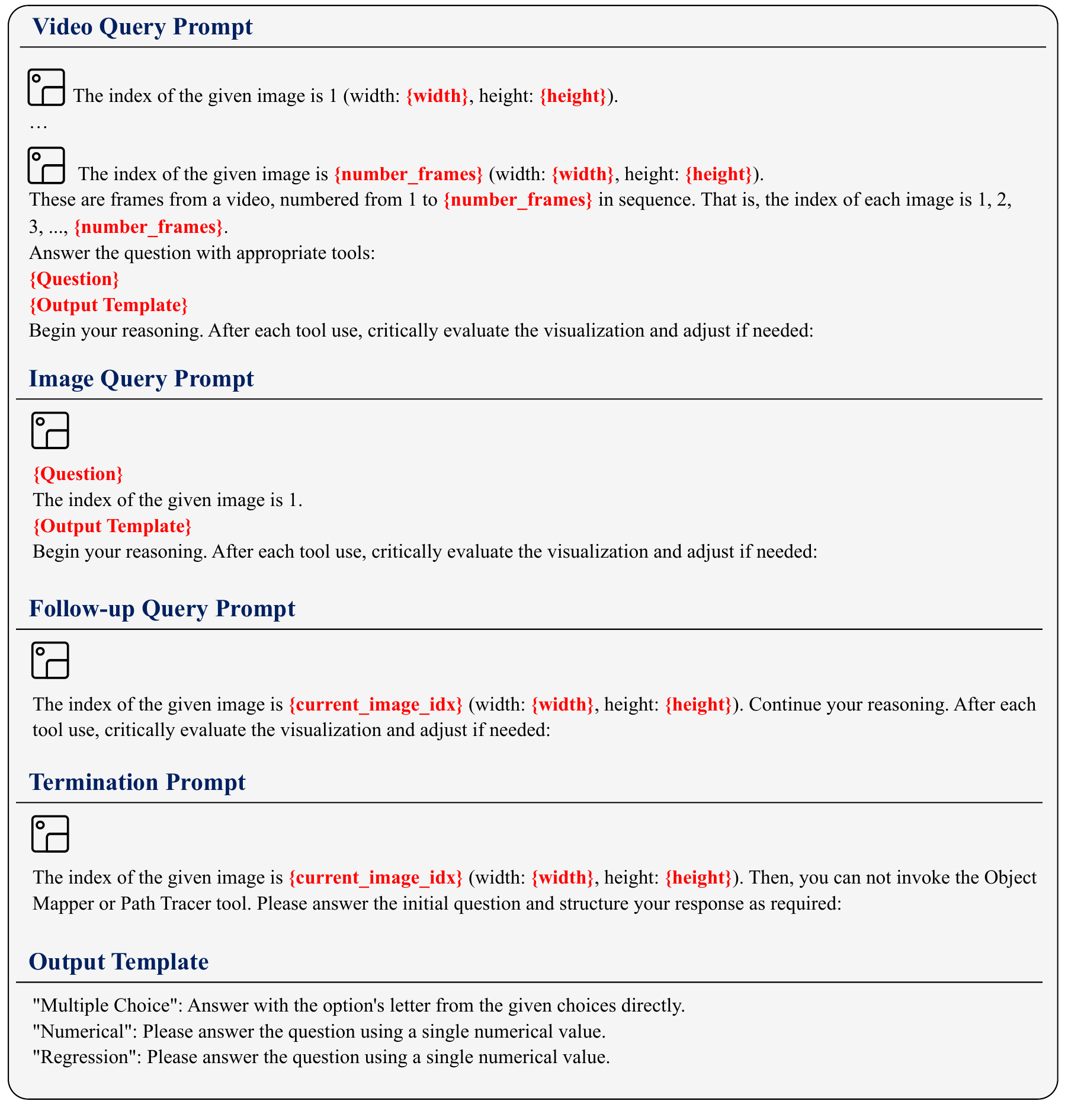}
  \vspace{-20pt}
  \caption{Query prompt and output template used in \textsc{ViLaSR}.}
  \label{fig:conv_prompt}
\end{figure}

In this section, we present the comprehensive prompt templates utilized in \textsc{ViLaSR}, including both the system prompt and user prompt, as illustrated in Figure~\ref{fig:sys_prompt} and Figure~\ref{fig:conv_prompt}, respectively.
We use these prompt templates for both generating reasoning paths in cold start data and inference of \textsc{ViLaSR}.

\subsection{System prompt of \textsc{ViLaSR}}
The system prompt utilized in the \textsc{ViLaSR} reasoning framework is presented in Figure~\ref{fig:sys_prompt}.

\subsection{Query prompt template of \textsc{ViLaSR}}
The query prompt utilized in the \textsc{ViLaSR} reasoning framework is presented in Figure~\ref{fig:conv_prompt}. Our framework utilizes three distinct types of prompts throughout the reasoning process:

\begin{itemize}
    \item \textbf{Initial Query Prompt} (Image/Video Query Prompt): Initiates the reasoning process by establishing the initial visual context and query.
    \item \textbf{Follow-up Query Prompt}: Guides subsequent reasoning steps by incorporating intermediate results and accumulated context.
    \item \textbf{Final Answer Prompt}: Terminates the reasoning process and enforces answer generation when either the maximum number of processed images or reasoning steps is reached.
\end{itemize}

\begin{figure}[!t]
  \centering
  \includegraphics[width=\linewidth]{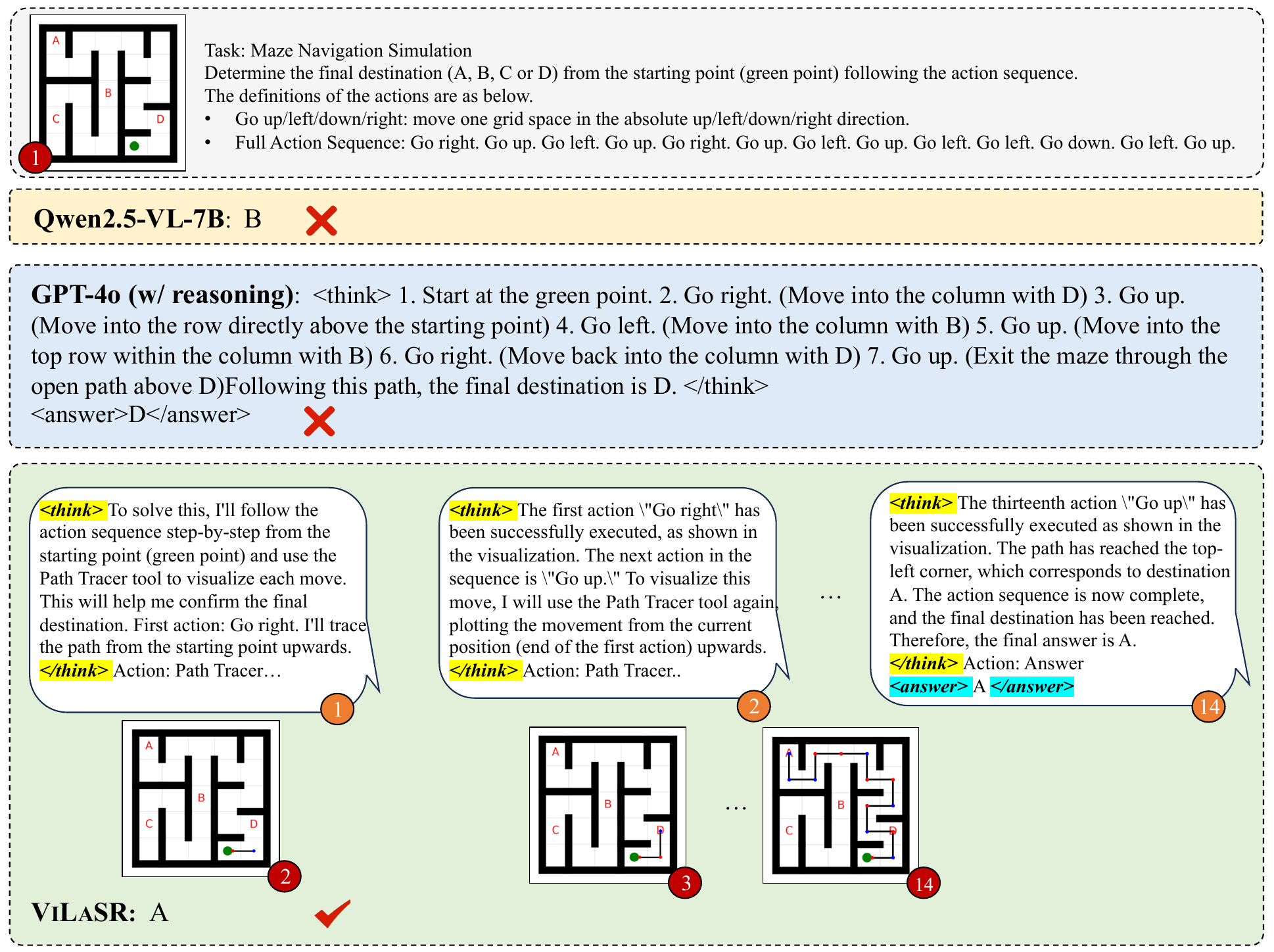}
  \vspace{-10pt}
  \caption{A visualization example of spatial reasoning approaches on the Maze benchmark, including Qwen-2.5-VL-7B, GPT-4o, and \textsc{ViLaSR}.}
  \label{fig:case_maze}
\end{figure}

% \begin{figure}[!t]
%   \centering
%   \includegraphics[width=\linewidth]{image/case study_embspatial.pdf}
%   \vspace{-20pt}
%   \caption{A visualization example of spatial reasoning approaches on the EmbSpatial benchmark, including Qwen-2.5-VL-7B, GPT-4o, and \textsc{ViLaSR}.\gnote{update model name}}
%   \label{fig:case_embspatial}
% \end{figure}

\begin{figure}[!t]
  \centering
  \includegraphics[width=\linewidth]{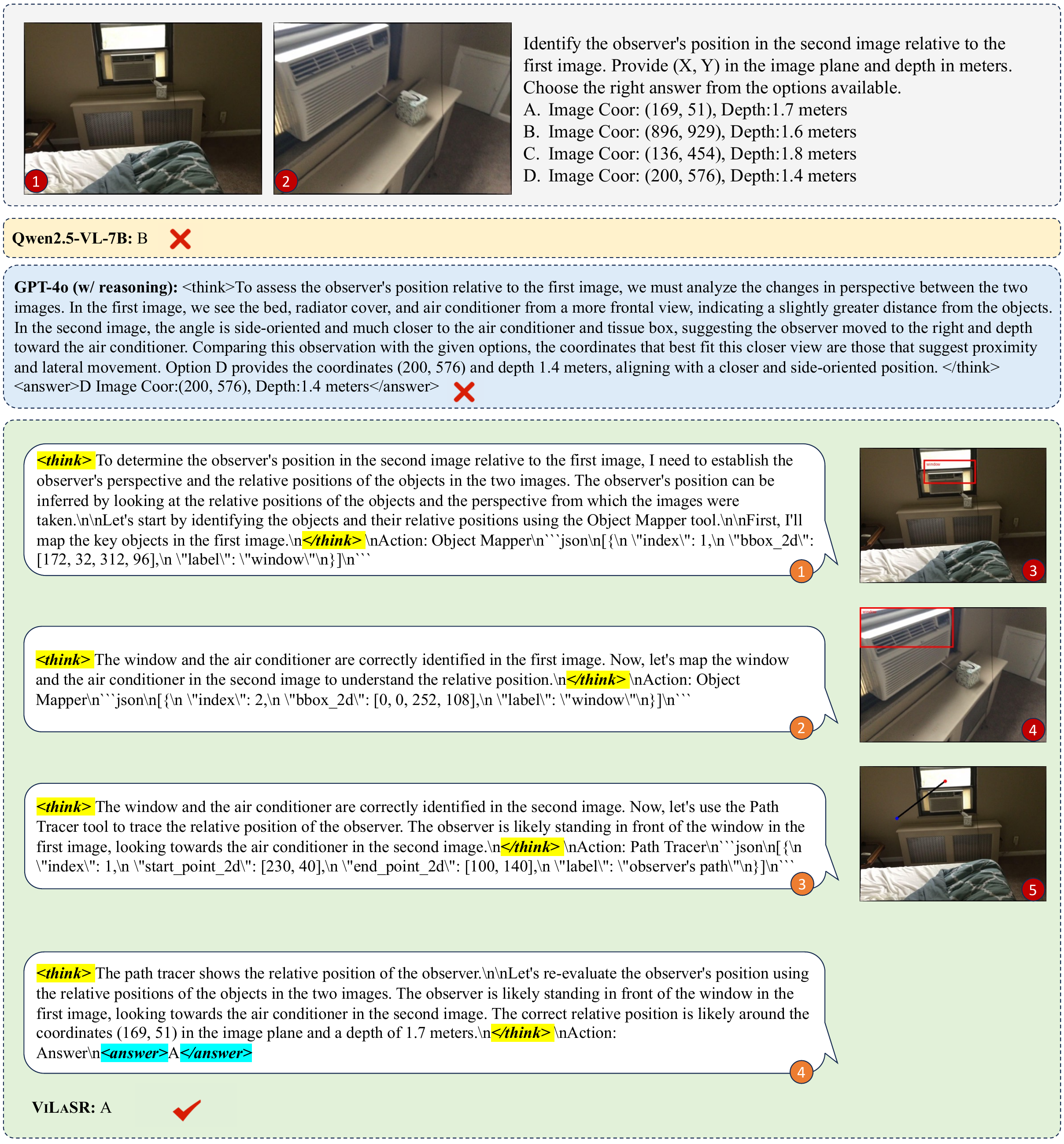}
  \vspace{-10pt}
  \caption{A visualization example of spatial reasoning approaches on the SPAR-Bench benchmark, including Qwen-2.5-VL-7B, GPT-4o, and \textsc{ViLaSR}.}
  \label{fig:case_spar}
\end{figure}

\begin{figure}[!t]
  \centering
  \includegraphics[width=\linewidth]{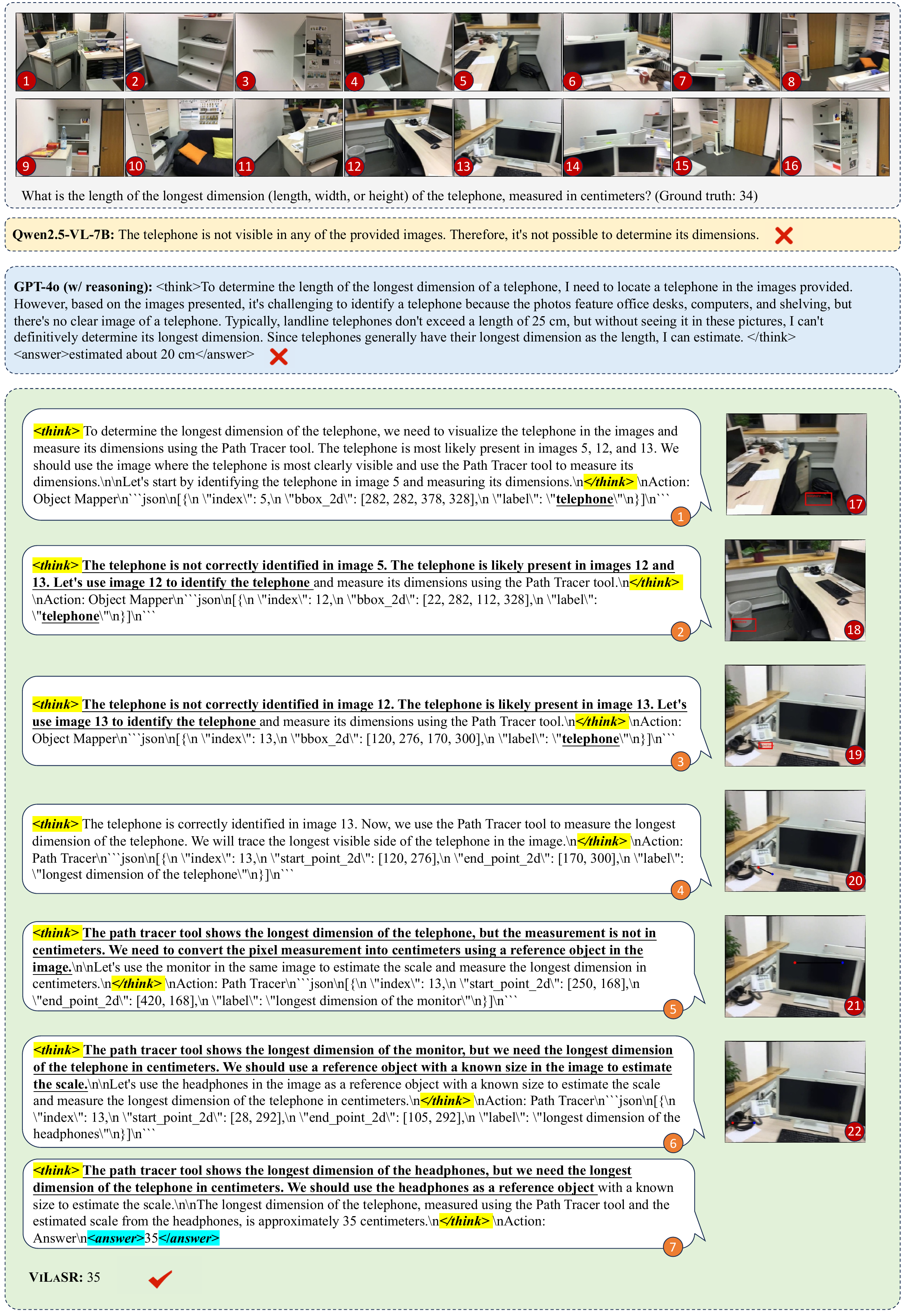}
  \vspace{-10pt}
  \caption{A visualization example of spatial reasoning approaches on the VSI-Bench benchmark, including Qwen-2.5-VL-7B, GPT-4o, and \textsc{ViLaSR}. We highlight the generated words of \textsc{ViLaSR} that exhibit reflection behavior in \textbf{\underline{bold}}.}
  \label{fig:case_vsibench}
\end{figure}

\section{Visualization results}
\label{appendix:visualization}
To illustrate the effectiveness of our approach, we present three representative examples from the MAZE, SPAR-Bench, and VSI-Bench benchmarks in Figures~\ref{fig:case_maze}, \ref{fig:case_spar}, and \ref{fig:case_vsibench}, respectively. Note that the ``Path Tracer'' tool refers to drawing auxiliary lines, and the ``Object Mapper'' tool refers to annotating bounding boxes.

In the maze navigation task (Figure~\ref{fig:case_maze}), while Qwen2.5-VL-7B provides incorrect answers directly and GPT-4o attempts textual reasoning but fails to accurately track spatial transitions, \textsc{ViLaSR} successfully decomposes the task into interpretable steps. By using the ``Path Tracer'' tool to visualize and verify each movement through auxiliary lines, our model ensures accurate navigation through the maze, leading to the correct destination.

For multi-view reasoning (Figure~\ref{fig:case_spar}), \textsc{ViLaSR} demonstrates sophisticated spatial understanding by first mapping key objects (``window,'' ``air conditioner'') in both images using ``Object Mapper'', then systematically analyzing their relative positions. Through careful verification using ``Path Tracer,'' it correctly determines the observer's position change, while both baseline models struggle with perspective transformations and make incorrect judgments about movement direction.

In the video spatial reasoning task (Figure~\ref{fig:case_vsibench}), \textsc{ViLaSR} exhibits strong reflection capability and systematic problem-solving. When initial attempts to locate and measure the telephone in images 5 and 12 fail, it self-corrects and identifies the correct object in image 13. Furthermore, it shows a sophisticated measurement strategy by using reference objects (``monitor,'' ``headphones'') to establish scale and convert pixel measurements to centimeters. In contrast, baseline models either fail to locate the target object (Qwen2.5-VL-7B) or make rough estimations without proper measurement (GPT-4o).

% Similarly, for real-world spatial relationship understanding (Figure~\ref{fig:case_embspatial}), \textsc{ViLaSR} demonstrates a systematic approach by first using ``Object Mapper'' to establish precise object locations by annotating bounding boxes, followed by ``Path Tracer'' to explicitly verify relative positions. This two-step visual reasoning process helps avoid ambiguity in spatial relationships, particularly in determining left-right relationships from the viewer's perspective. While GPT-4o reaches the correct conclusion through textual reasoning, our visual approach provides more explicit and verifiable reasoning steps.

% For video-based spatial reasoning  (Figure~\ref{fig:case_vsibench}), \textsc{ViLaSR} further demonstrates its advantage in handling complex multi-frame scenarios. When tasked with determining relative distances between objects across multiple viewpoints, \textsc{ViLaSR} employs a systematic approach: first using ``Object Mapper'' to identify objects from multiple frames, then applying ``Path Tracer'' to measure and compare distances. This approach enables accurate spatial comparison even when object positions and camera angles change across frames. In contrast, Qwen2.5-VL-7B struggles to effectively process lengthy video sequences, as it lacks the ability to selectively encode and analyze key frames, leading to information overload and incorrect spatial reasoning. This case highlights the importance of structured visual operations in managing and analyzing complex temporal-spatial information.

These cases illustrate how drawing operations enable more reliable spatial reasoning by grounding abstract relationships in concrete visual representations.

% Visualization results will also be presented in the subsequent technical appendices along with the supplementary material.\gnote{update. We had better to cherry-pick some examples exhibiting reflection.}

% \section{Case study}

\section{Complexity analysis}\label{complexity}
To thoroughly assess the computational efficiency of \textsc{ViLaSR}, we present a comprehensive analysis comparing our approach with the base model Qwen2.5-VL-7B model. This analysis encompasses both theoretical complexity bounds and empirical resource utilization, providing insights into the framework's scalability and computational characteristics.

\subsection{Theoretical time complexity}
We analyze the computational complexity across three model variants: the base model (performing direct answer generation), standard reasoning models (generating reasoning chains followed by answer derivation), and our \textsc{ViLaSR} model (employing iterative visual drawing and thinking). Let $M$ denote the input length (including both image and text tokens), $N$ the answer length, $L$ the per-step reasoning path length, and $S$ the number of reasoning steps, where $S=1$ for both base model and standard reasoning approaches. Empirically, \textsc{ViLaSR} achieves shorter per-step reasoning paths (smaller $L$) compared to single-step approaches due to its decomposition of reasoning into interpretable visual operations. Table~\ref{tab:complexity} presents a comparative analysis of their time complexities.

\begin{table}[!t]
    \centering
    \caption{Complexity analysis of different reasoning approaches.}
    \label{tab:complexity}
    \renewcommand{\arraystretch}{1.3}
    \begin{tabular}{llll}
        \hline
        \textbf{Model Type} & \textbf{Input Length} & \textbf{Output Length} & \textbf{Time Complexity} \\
        \hline
        Base Model & $M$ & $N$ & $\mathcal{O}(M \cdot N)$ \\
        Standard Reasoning & $M$ & $L + N$ & $\mathcal{O}(M \cdot (L + N))$ \\
        \midrule
        \textsc{ViLaSR} (Ours) & $M + \sum_{i=1}^{S}M_i$ & $S \cdot L + N$ & \makecell[l]{
            $\mathcal{O}\big(S^2 \cdot L \cdot (L + M)$\\
            $+ S \cdot (L + M)(L + N)\big)$} \\
        \hline
    \end{tabular}
\end{table}

The total computational complexity is:
\begin{equation}
T = \sum_{k=1}^{S-1} [(M + \sum_{i=1}^{k-1}(L_i + M_i)) \cdot L_k] + (M + \sum_{i=1}^{S-1}(L_i + M_i)) \cdot (L_S + N),
\end{equation}
where $L_k$ is the length of the reasoning path generated at step $k$, and $M_k$ is the additional multimodal context incorporated at step $k$.

Assuming each reasoning step has approximately equal length, i.e., $L_i \approx L$, $M_i \approx M$ for all $i$, we can approximate:
\begin{equation}
    T = \mathcal{O}\left(S^2 \cdot L \cdot (L + M) + S \cdot (L + M)(L + N)\right).
\end{equation}

\subsection{Practical runtime}
To provide empirical evidence for our complexity analysis, we evaluate \textsc{ViLaSR}'s runtime performance across different benchmarks. Table~\ref{tab:runtime} illustrates the average runtime time and the average number of reasoning steps per sample.
\begin{table}[!t]
    \centering
    \caption{Average runtime and reasoning steps of \textsc{ViLaSR} across different benchmarks.}
    \label{tab:runtime}
    \renewcommand{\arraystretch}{1.3}
    \begin{tabular}{lcc}
        \hline
        \textbf{Benchmark} & \textbf{Avg. Runtime (s/sample)} & \textbf{Avg. Reasoning Steps} \\
        \hline
        Maze         & 4.4  & 10.8  \\
        EmbSpatial         &  1.2  & 3.1 \\
        VSIbench       & 7.8   & 4.5 \\
        % Spar-Bench   & -   & - \\
        \hline
    \end{tabular}
\end{table}

\section{Limitations}~\label{limitations}

While our work demonstrates promising results in spatial reasoning tasks, several limitations warrant discussion:

First, our training framework primarily focuses on multiple-choice and numerical questions due to their amenability to automated evaluation, excluding more complex spatial reasoning scenarios that require free-form textual descriptions (e.g., detailed motion trajectory analysis). This limitation constrains the model's ability to handle open-ended spatial reasoning tasks.

Second, the effectiveness of our drawing operations is inherently limited by the 2D nature of the visual interface. Complex 3D spatial relationships and viewpoint changes, which are common in real-world scenarios, may not be adequately captured by our current drawing primitives. This limitation particularly affects the model's performance on tasks involving 3D object relationships or dynamic camera movements.

Finally, the computational cost of our three-stage training pipeline, especially during the reinforcement learning stage, may limit its accessibility to researchers with limited computational resources. 

\section{Broader impact}\label{broader_impact}

Our work on enhancing spatial reasoning capabilities in vision-language models has both positive and potential negative societal implications that warrant careful consideration.

On the positive side, improved spatial reasoning capabilities could significantly benefit various applications in robotics, autonomous navigation, and assistive technologies. For instance, more accurate spatial understanding could help robots better navigate complex environments and assist people with visual impairments in daily tasks. In educational settings, these models could provide interactive tools for teaching spatial concepts and geometric reasoning. Additionally, the interpretable nature of our drawing-based reasoning approach enhances model transparency, potentially increasing trust and adoption in critical applications.

However, several potential negative impacts need to be addressed. First, the enhanced spatial reasoning capabilities could be misused for surveillance purposes, enabling more sophisticated tracking and monitoring systems that could infringe on privacy rights. Second, there might be accessibility issues as the drawing-based reasoning approach assumes users have access to and can interact with visual interfaces, potentially excluding certain user groups.

To mitigate these concerns, we recommend: (1) implementing strict usage guidelines and access controls when deploying these models in sensitive applications, (2) exploring alternative interaction modalities to make the technology more accessible. We also encourage future research to focus on developing privacy-preserving spatial reasoning techniques.

To ensure responsible deployment of our technology, we have implemented several safeguards in our release. First, our model access will be gated through an API that requires user agreement to usage guidelines, specifically prohibiting applications in surveillance or privacy-invasive systems. Second, we provide detailed documentation about the model's capabilities and limitations, along with best practices for responsible implementation in different application scenarios.

\end{document}